\UseRawInputEncoding
\documentclass[journal]{IEEEtran}

\usepackage[numbers,sort&compress]{natbib}
\usepackage[colorlinks=true, linkcolor=blue, citecolor=blue, urlcolor=blue]{hyperref}
\usepackage{graphicx,float,amsmath,amssymb,pifont,epstopdf,array,makecell}
\usepackage[switch]{lineno}
\usepackage{color,subeqnarray,cases}
\usepackage{subfigure}
\usepackage{algorithm}
\usepackage{algpseudocode}
\usepackage{caption}

\begin{document}

\title{\huge{Sum Rate Maximization in STAR-RIS-UAV-Assisted Networks: A CA-DDPG Approach for Joint Optimization}
}

\author{Yujie~Huang, 
        Haibin~Wan, 
        Xiangcheng~Li,
        Tuanfa~Qin,
        Yun Li,~\IEEEmembership{Member,~IEEE},
        Jun Li,~\IEEEmembership{Fellow,~IEEE},
        and Wen Chen,~\IEEEmembership{Senior Member,~IEEE}

\thanks{Yujie Huang is with the School of Computer and Electronic Information, Guangxi University, Nanning, China (e-mail: 2313391011@st.gxu.edu.cn).}

\thanks{Haibin Wan is with the School of Computer and Electronic Information, Guangxi University, Nanning, China (e-mail: hbwan@gxu.edu.cn).}

\thanks{Xiangcheng Li is with the School of Computer and Electronic Information, Guangxi University, Nanning, China (e-mail: xcli@gxu.edu.cn).}

\thanks{Tuanfa Qin is with the School of Computer and Electronic Information, Guangxi University, Nanning, China (e-mail: tfqin@gxu.edu.cn).}

\thanks{Yun Li is with the College of Mathematics and Computer Science, Zhejiang A\&F University, Hangzhou, China (e-mail: liyun@zafu.edu.cn).}

\thanks{Jun Li is with the School of Information Science and Engineering, Southeast University, Nanjing, China (e-mail: jun.li@seu.edu.cn).}

\thanks{Wen Chen is with the Department of Electronic Engineering Shanghai Jiao Tong University, Shanghai, China (e-mail: wenchen@sjtu.edu.cn).}
}

\maketitle

\begin{abstract}
With the rapid advances in programmable materials, reconfigurable intelligent surfaces (RIS) have become a pivotal technology for future wireless communications. The simultaneous transmitting and reflecting reconfigurable intelligent surfaces (STAR-RIS) can both transmit and reflect signals, enabling comprehensive signal control and expanding application scenarios. This paper introduces an unmanned aerial vehicle (UAV) to further enhance system flexibility and proposes an optimization design for the spectrum efficiency of the STAR-RIS-UAV-assisted wireless communication system. We present a deep reinforcement learning (DRL) algorithm capable of iteratively optimizing beamforming, phase shifts, and UAV positioning to maximize the system’s sum rate through continuous interactions with the environment. To improve exploration in deterministic policies, we introduce a stochastic perturbation factor, which enhances exploration capabilities. As exploration is strengthened, the algorithm’s ability to accurately evaluate the state-action value function becomes critical. Thus, based on the deep deterministic policy gradient (DDPG) algorithm, we propose a convolution-augmented deep deterministic policy gradient (CA-DDPG) algorithm that balances exploration and evaluation to improve the system’s sum rate. The simulation results demonstrate that the CA-DDPG algorithm effectively interacts with the environment, optimizing the beamforming matrix, phase shift matrix, and UAV location, thereby improving system capacity and achieving better performance than other algorithms.
\end{abstract}

\begin{IEEEkeywords}
Reconfigurable intelligent surface, Unmanned aerial vehicle, Deep reinforcement learning, Convolutional neural network, Beamforming.
\end{IEEEkeywords}

\section{INTRODUCTION}
\IEEEPARstart{I}{n} recent years, with the widespread deployment of the fifth-generation (5G) mobile communication networks and the continuous advancement of related software and hardware technologies, 5G has been able to provide users with higher reliability, enhanced security, faster data transmission speeds, and lower latency. Meanwhile, the growing demand for communication between various devices has further driven the development of 5G, while also placing higher demands on the future of wireless communications \cite{9999288,10158439,10387440,8792139,9548837}.

As the successor to 5G, the sixth-generation (6G) mobile communication network is expected to achieve higher data rates, expanded coverage, and robust security. However, the continuous advancement of wireless communication has led to the use of increasingly higher transmission frequencies, inevitably introducing inherent challenges such as susceptibility to interference, reduced coverage, weakened penetration capability, and related issues. Moreover, the explosive growth in data volume and the increasing complexity of hardware devices have created significant cost challenges. These issues present a critical dilemma for the future development of wireless communication: how to enhance system performance while minimizing costs and resource consumption has become one of the core problems requiring urgent resolution \cite{9475160,8869705,10106261,10054381}.

Benefiting from collaborative advancements across multiple disciplines, the challenge of high-frequency signal blockage by large obstacles has been effectively mitigated through the advent of RIS \cite{8910627}. RIS employs a class of metamaterials termed ``information metamaterials'', comprising a planar array of passive electromagnetic reflective elements that are dynamically controlled via software or hardware configurations. The core functionality of RIS lies in dynamically adjusting the reflection phase shifts and amplitudes of incident signals to steer directional beams \cite{8981888}. By harnessing real-time channel state information (CSI) from signal propagation feedback, RIS enables adaptive wireless environment reconfiguration through software-defined control. This not only introduces new spatial degrees of freedom to enhance link performance but also lays the foundation for intelligent and programmable radio environments \cite{9110889}. Beyond signal reflection capabilities, RIS is characterized by its compact form factor and low cost, which significantly simplifies the deployment of communication infrastructure. These advantages enable RIS to be flexibly installed on building exteriors, indoor walls, vehicles, UAVs, and virtually any location due to its miniaturized design. Furthermore, the cost-effectiveness of RIS enables large-scale deployment without incurring prohibitive expenses \cite{9424177}.

The proposal of STAR-RIS represents an expansion in the spatial dimension. Compared with reflective RIS, this form not only retains the signal reflection capability of reflective RIS but also caters to users and devices on the back of the RIS. This feature enables it to meet the communication needs of both sides of the RIS simultaneously, making it applicable in a wider range of scenarios and more effective in solving communication problems in complex environments.

RIS has indeed improved transmission performance, but has also introduced complex optimization problems. In recent years, extensive research has focused on optimizing beamforming and phase shifts to achieve optimal spectral efficiency. Due to the inherent complexity of RIS optimization problems, finding a closed-form solution is often difficult. Additionally, as the number of RIS reflective elements increases, costs can rise uncontrollably. Therefore, selecting an appropriate method has become a significant challenge in addressing RIS-related beamforming and phase shift optimization problems \cite{Cao2025}.

Over the past few years, algorithms in artificial intelligence (AI) and machine learning (ML) have remained a focal point of academic research. Notably, some DRL algorithms have demonstrated remarkable effectiveness in solving optimization problems in communication systems \cite{10361836,10100477}. By learning optimal behavioral policies through agent-environment interactions, some DRL algorithms are particularly well-suited for stochastic wireless communication environments \cite{9729826}. This paper assumes that direct transmissions between the base station (BS) and users are entirely blocked. Due to multi-user interference, the optimization problem is non-convex, with no known optimal solution. To address this problem, we propose a multi-user multiple-input single-output (MU-MISO) downlink transmission algorithm based on DDPG, designed to jointly optimize beamforming, phase shifts, and the location of UAV to maximize the sum rate.

\subsection{Prior Works}
The joint optimization of beamforming and phase shifts in RIS-assisted wireless communication systems has been extensively studied. In \cite{9032163}, the joint optimization problem in MU-MISO millimeter-wave communication systems was addressed using the gradient-projection (GP) method. In \cite{9912342}, rate-splitting multiple access (RSMA) technology was explored in an RIS-assisted uplink system to enhance the performance of the sum rate. In \cite{9139465}, constructive interference (CI) was leveraged to design RIS phase shifts, enabling direct-link and reflected signals to combine at each user's side. The proposed algorithm was used to alternately optimize the precoder and RIS reflection elements to minimize the symbol error rate (SER) in \cite{9097454}. In \cite{8982186}, under both perfect and imperfect CSI conditions, a low-complexity block coordinate descent (BCD) method was proposed to decompose the problem into subproblems for precoding and phase shift optimization.

Due to RIS's superior performance, its integration with diverse technologies and environments has become a crucial research direction. Zhang \textit{et al.} \cite{9459505} combined RIS with cell-free networks to enhance communication capacity by mitigating inter-cell interference. Nassirpour \textit{et al.} \cite{10220203} established centralized and distributed RIS to evaluate the performance of RIS-assisted Internet of Things (IoT) networks and the optimal location distribution of RIS. Niu \textit{et al.} \cite{9903846} explored the use of RIS to improve the security performance of IoT network communication. In \cite{10494543}, the integration of UAVs with compact and portable RIS was proposed to address communication challenges in special scenarios and enhance security in multi-user network communication. In \cite{9996966}, power allocation, RIS passive beamforming, and UAV trajectories were optimized for the worst-case downlink secrecy rate in vehicular networks with eavesdroppers. Lin \textit{et al.} \cite{10726590} focused on green communication in RIS-assisted UAV systems and investigated the energy efficiency optimization problem for fixed-wing UAV communications under RIS assistance. For integrated sensing and communication (ISAC), Luo \textit{et al.} \cite{10052711} proposed an alternating optimization (AO) algorithm combining majorization-minimization (MM), penalty-based, and manifold methods to solve non-convex problems in RIS-assisted ISAC systems. Wang \textit{et al.} \cite{10056867} analyzed average mean-square error (MSE) in practical RIS-MIMO systems with hardware impairments, phase noise, and imperfect CSI. STAR-RIS, as a revolutionary concept, has attracted significant attention due to its capability to achieve full-space signal coverage, substantially expanding the application boundaries and design flexibility of RIS systems. Mu \textit{et al.} \cite{9570143} systematically proposed the fundamental signal model for STAR-RIS and innovatively designed three practical operating protocols: Energy Splitting (ES), Mode Switching (MS), and Time Switching (TS). Zuo \textit{et al.} \cite{9863732} further integrated STAR-RIS with NOMA and proposed the STAR-RIS-NOMA system. Du \textit{et al.} \cite{10086660} explored the application of STAR-RIS in wireless-powered IoT networks.

ML has been a research hotspot in the scientific community in recent years, especially since many unsupervised ML algorithms are well-suited for solving complex, variable, high-dimensional problems in wireless communication systems \cite{9210812}. Extensive research has demonstrated that ML, as one of the most powerful AI tools, meets the demands of next-generation mobile networks in terms of speed, intelligence, and applications \cite{7792374,DAS2023109581}. In \cite{9627230}, the research gap was addressed through a comparison of the performance of different ML tools in wireless networks. Deep learning (DL) and reinforcement learning (RL) algorithms were employed to optimize RIS-assisted non-orthogonal multiple access (NOMA) and orthogonal multiple access (OMA) systems, respectively. In \cite{10193812}, physical layer security in RIS-assisted ISAC systems was investigated, where artificial noise (AN) was employed to disrupt eavesdroppers. Xu \textit{et al.} \cite{10685526} proposed a parameterized deep Q-network (PG-PDQN) algorithm to tackle the challenges of mixed discrete-continuous action spaces in RIS-assisted IoT systems. To maximize system capacity under energy constraints in RIS-assisted UAV NOMA systems, Zhang \textit{et al.} \cite{9919620} employed a double deep Q-network (DDQN)-based DRL algorithm to jointly optimize UAV flight trajectories and RIS phase shifts. In \cite{9967967}, the RIS-assisted HSR network was investigated to address QoS degradation in high-speed railway (HSR) networks caused by external interference.

Online DRL algorithms are pivotal for future communications, primarily because they can operate without labeled datasets or offline training. In particular, the DDPG algorithm has demonstrated exceptional performance in solving RIS-assisted wireless communication problems. Consequently, researchers have extensively investigated DDPG-based algorithms for RIS-assisted wireless communication systems. Huang \textit{et al.} \cite{9110869} applied DDPG to optimize precoding and phase shift matrices in RIS-assisted MU-MISO system to maximize rewards. In \cite{10692454}, a distributed downlink beamforming method was proposed, which utilized the distributed deterministic policy gradient with prioritized experience replay (PRE-D4PG) algorithm to enhance system performance. Li \textit{et al.} \cite{10261216} presented a distributed training decentralized execution (DTDE) algorithm based on DDPG, achieving performance gains in RIS-assisted system. Faisal \textit{et al.} \cite{10637275} introduced the twin-delayed DDPG (TD3) algorithm to mitigate $Q$ value overestimation bias. In \cite{10504542}, the TD3 algorithm was employed to optimize RIS phase shifts and energy allocation for IoT devices (IoTDs). In \cite{10603314}, long short-term memory (LSTM) networks were integrated with DDPG to control data delivery. Similarly, Wu \textit{et al.} \cite{10288083} proposed an adaptive learning DDPG (AL-DDPG) algorithm for the joint optimization of RIS phase shifts and BS beamforming in satellite-aerial-ground integrated networks (SAGINs). Guo \textit{et al.} \cite{10109153} designed a multi-objective DDPG (MO-DDPG) algorithm to maximize the achievable sum rate under multi-objective optimization frameworks.

\subsection{Contributions}
To the best of our knowledge, the sum rate optimization problem in STAR-RIS-UAV systems has not been thoroughly investigated using DRL algorithms. In this paper, we propose a STAR-RIS-UAV-assisted multi-user system model. To maximize the sum rate of users, we jointly optimize BS beamforming, RIS phase shifts, and the location of the UAV using a DRL algorithm. Assuming complete blockage between the BS and users, we address the intractable non-convex downlink optimization problem, whose optimal solution is unknown, by developing a novel DRL-based algorithm. Specifically, we propose a CA-DDPG algorithm derived from DDPG to solve this problem. Compared to conventional DDPG, our method achieves superior performance. The main contributions are summarized as follows:
\begin{itemize}
    \item The installation of STAR-RIS can effectively bypass obstacles to enhance the speed and quality of communication. Additionally, the integration of UAVs further increases the flexibility of STAR-RIS placement. By mounting STAR-RIS on a UAV, its spatial position can be dynamically adjusted to identify favorable channel conditions, thereby facilitating more efficient signal transmission. To optimize the sum rate of users, we establish a downlink transmission scenario in a STAR-RIS-UAV-assisted multi-user system, where the sum rate is maximized by jointly optimizing BS beamforming, RIS phase shifts, and the location of the UAV.
    \item Considering the advantages of DRL algorithms, we design a new DRL algorithm called CA-DDPG based on the DDPG framework to solve the proposed non-convex optimization problem. To enhance the exploration of actions in deterministic policies, we introduce a stochastic perturbation factor. Additionally, the original evaluation network limits further improvements in rewards. Therefore, we extend the original DDPG framework by designing a more complex critic network with higher evaluation capabilities. Specifically, inspired by the ability of convolutional neural networks (CNNs) to extract local features, we incorporate convolutional layers into the critic network to improve its evaluation performance.
    \item To validate CA-DDPG, we conduct numerical simulations against multiple algorithms, including the traditional DDPG algorithm, TD3, and DDPG-based DTDE. Specifically, we configure multiple parameter sets with varying environmental configurations, including transmit power, number of BS antennas, number of users, number of RIS elements, learning rate, and decay coefficient. In addition, we study the complexity and convergence of the algorithm and investigate the performance of the CA-DDPG compared to various algorithms under imperfect CSI.
\end{itemize}

The rest of this paper is organized as follows. Section II presents the system model and the sum rate maximization problem. The details of the CA-DDPG algorithm for joint beamforming, phase shifts, and UAV location design are described in Section III. Section IV provides simulation results and analysis. Finally, conclusions are presented in Section V.

Notations: For matrix $\mathbf{H}$, $\mathbf{H}(i,j)$ denotes the entry at the $i$-th row and the $j$-th column. $\mathbf{H}^{T}$ and $\mathbf{H}^{H}$ represent transpose and conjugate transpose of matrix $\mathbf{H}$, and $\mathbf{h}_{k}$ is the $k$-th column vector of $\mathbf{H}$. For column vector $\mathbf{x}$, $\mathbf{x}(i)$ is the $i$-th element, and $\|\mathbf{x}\|$ denotes its magnitude. $\operatorname{diag}(\mathbf{x})$ represents a diagonal matrix, whose diagonal elements are the corresponding elements of vector $\mathbf{x}$. $ \mathbb{C}^{N \times 1} $  represents a complex vector with dimension $N$, and $ \mathbb{C}^{N \times M} $ represents a matrix in the complex field. $ \mathrm{E}[\cdot] $ is statistical expectation, and $\mathcal{CN}(0, \sigma^2)$ denotes a complex Gaussian distribution with mean 0 and variance $\sigma^2$. For complex number $c$, $|{c}|$ denotes the absolute value of the complex number. $Re(c)$ and $Im(c)$ denote the real and imaginary parts of $c$, respectively.

\section{SYSTEM MODEL AND PROBLEM FORMULATION}
\subsection{System Model}

\begin{figure}[t]
\centering
\includegraphics[width=3.5in]{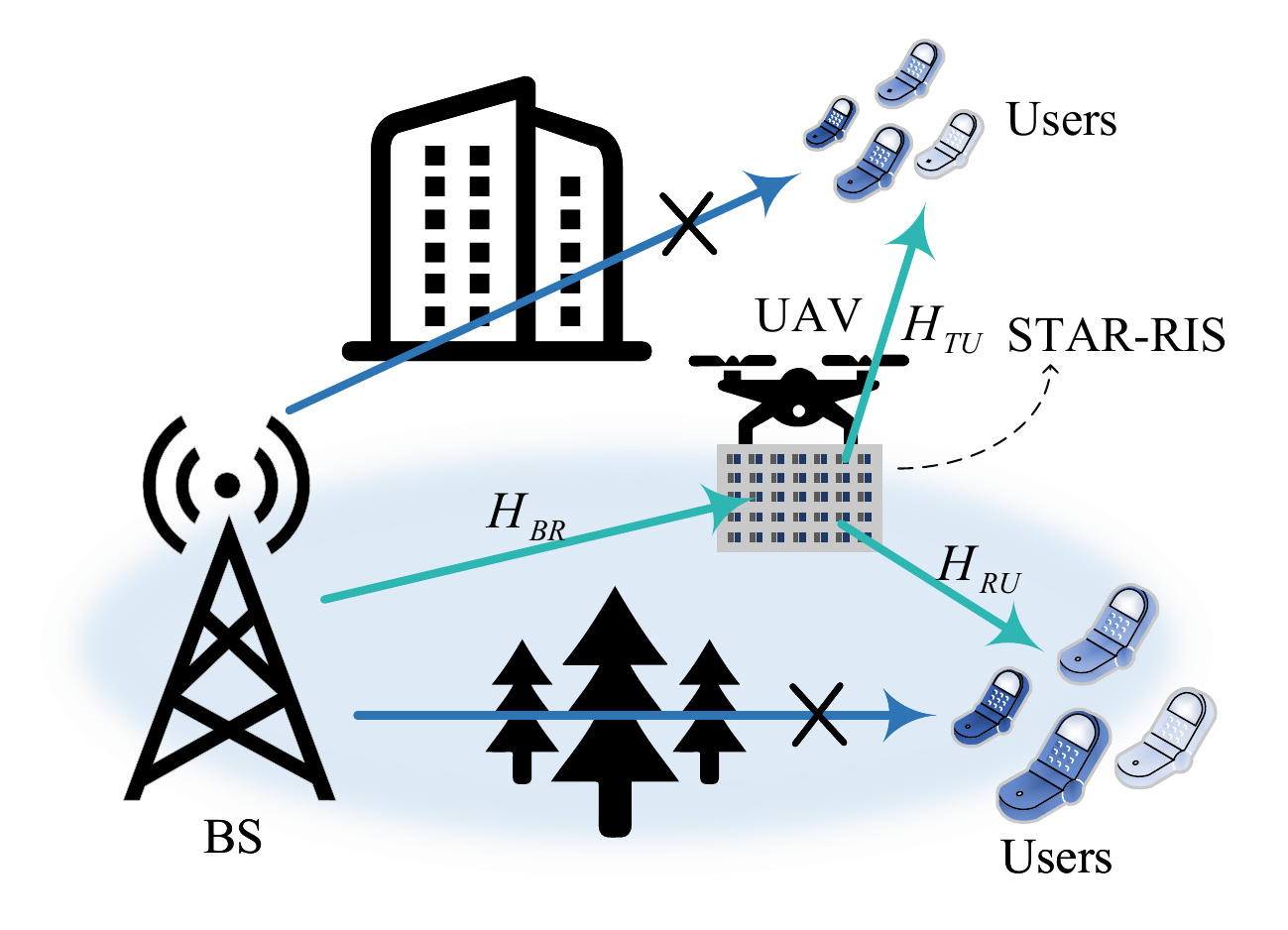}
\caption{STAR-RIS-UAV communication system.}
\label{fig.1}
\end{figure}

We consider an STAR-RIS-UAV-assisted multi-user downlink wireless communication system, consisting of a BS with $M$ antennas, an STAR-RIS with $N$ reflective elements and $N$ transmissive elements, $K$ single-antenna users on the reflection side and another $K$ single antenna users on the transmission side, as illustrated in Fig.\ref{fig.1}. Due to large obstacles between the BS and users in practical scenarios, the direct link is completely blocked, leading to severe degradation or even failure of transmission quality. To address this, the STAR-RIS is deployed to reconstruct the communication path by intelligently reflecting signals. Specifically, signals transmitted by the BS are phase-adjusted via the controller-regulated RIS and then reflected or transmitted to the users. The channel matrices from the BS to the RIS, the channel vector from the RIS to the users on the reflective side, and the users on the transmissive side are denoted as $\mathbf{H}_{BR}$ $\in$ $\mathbb{C}^{N\times M}$, $\mathbf{H}_{RU}$ $\in$ $\mathbb{C}^{N\times 1}$ and $\mathbf{H}_{TU}$ $\in$ $\mathbb{C}^{N\times 1}$ respectively. The entries of $\mathbf{H}_{BR}$,  $\mathbf{H}_{RU}$ and $\mathbf{H}_{TU}$ are modeled as independent and identically distributed (i.i.d.) complex Gaussian random variables. The BS is located at horizontal coordinates $\mathbf{C}_{B}=[x,y]$ with height ${h}_{B}$. The UAV is equipped with a STAR-RIS and continuously moves in the air; its position at time $t$ is denoted as $\mathbf{C}_{R}[t]=[x[t],y[t]]$, and it flies at a fixed altitude ${h}_{R}$. Users on the reflection side are located at coordinates $\mathbf{C}_{R,K}=[x_{R,K},y_{R,K}]$, while those on the transmission side are located at coordinates $\mathbf{C}_{T,K}=[x_{T,K},y_{T,K}]$.

The signal received by the $k$-th user on the reflective surface and the signal received by the $k$-th user on the transmissive surface can be represented as
\begin{align}
\label{equ_1}
y_{r,k} &= \mathbf{H}_{R U}^{T} \mathbf{\Theta_{R}} \mathbf{H}_{B R} \mathbf{W}_{r} \mathbf{x}_{r} + z_{r,k},
 \\
\label{equ_2}
y_{t,k} &= \mathbf{H}_{T U}^{T} \mathbf{\Theta_{T}} \mathbf{H}_{B R} \mathbf{W}_{t} \mathbf{x}_{t} + z_{t,k},
\end{align}
where $y_{r,k}$, $y_{t,k}$ denote the signals received by the $k$-th user on the reflective and transmissive side. The signals are represented by $\mathbf{x}_{r}$, $\mathbf{x}_{t}$ and satisfy $\mathbb{E}[\mathbf{x}_{r}^2] = 1$, $\mathbb{E}[\mathbf{x}_{t}^2] = 1$. $z_{r,k}$, $z_{t,k}$ represent zero-mean additive white Gaussian noise (AWGN) with variance $\sigma^2$, i.e., $z_{r,k}  \sim \mathcal{CN}(0, \sigma^2)$ and $z_{r,k}  \sim \mathcal{CN}(0, \sigma^2)$. $\mathbf{W}_{r} \in \mathbb{C}^{M \times K}$, $\mathbf{W}_{t} \in \mathbb{C}^{M \times K}$ denote the BS beamforming matrix on the reflective and transmissive side, respectively. The phase shift matrices of the reflective RIS and transmissive RIS are defined as
\begin{align}
\label{equ_3}
\mathbf{\Theta}_{R} = \text{diag}[\sqrt{\beta_r^1}\phi_{r,1}, \sqrt{\beta_r^2}\phi_{r,2}, \ldots, \sqrt{\beta_r^i}\phi_{r,i}, \ldots, \sqrt{\beta_r^N}\phi_{r,N}],
 \\
\label{equ_4}
\mathbf{\Theta}_{T} = \text{diag}[\sqrt{\beta_t^1}\phi_{t,1}, \sqrt{\beta_t^2}\phi_{t,2}, \ldots, \sqrt{\beta_t^i}\phi_{t,i}, \ldots, \sqrt{\beta_t^N}\phi_{t,N}],
\end{align}
where $\phi_{r,i} = e^{j\varphi_{r,i}}$ and $\phi_{t,i} = e^{j\varphi_{t,i}}$. $\varphi_{r,i}$, $\varphi_{t,i}$ represent the RIS phase shift angles and satisfy $\varphi_{r,i} \in [0,2\pi)$, $\varphi_{t,i} \in [0,2\pi)$. $\beta_r^i$, $\beta_t^i$ indicate the STAR-RIS operating in the energy splitting mode, ensure each element operates in a mode that enables simultaneous reflection and transmission. $\beta_r^i \in [0,1]$, $\beta_t^i \in [0,1]$ and satisfy $\beta_r^i +\beta_t^i=1$. Note that $\mathbf{\Theta}=[\mathbf{\Theta_{r}},\mathbf{\Theta_{t}}]$ is a diagonal matrix whose diagonal elements are given by $\mathbf{\Theta_{r}}(i,i)=\phi_{r,i}$, $\mathbf{\Theta_{t}}(i,i)=\phi_{t,i}$ and satisfy $\left|\mathbf{\Theta_{r}}(i,i)\right|^2=1$, $\left|\mathbf{\Theta_{t}}(i,i)\right|^2=1$, where $\phi_{r,i}$, $\phi_{t,i}$ are the phase shift caused by the elements of the RIS.

We assume that the signal can be perfectly reflected and transmitted by the STAR-RIS without any signal loss. For convenience, the received signal models in (\ref{equ_1}) and (\ref{equ_2}) can be further rewritten as
\begin{equation}
\begin{split}
y_{r,k} = &\mathbf{H}_{RU}^T \mathbf{\Theta}_{R} \mathbf{H}_{BR} \mathbf{w}_{r,k} x_{r,k} \\
          &+ \sum_{n, n \neq k}^K \mathbf{H}_{RU}^T \mathbf{\Theta_{R} H}_{BR} \mathbf{w}_{r,n} x_{r,n} + z_{r,k},
\end{split}
\label{equ_5}
\end{equation}
\begin{equation}
\begin{split}
y_{t,k} = &\mathbf{H}_{TU}^T \mathbf{\Theta}_{T} \mathbf{H}_{BR} \mathbf{w}_{t,k} x_{t,k} \\
          &+ \sum_{n, n \neq k}^K \mathbf{H}_{TU}^T \mathbf{\Theta_{T} H}_{BR} \mathbf{w}_{t,n} x_{t,n} + z_{t,k},
\end{split}
\label{equ_6}
\end{equation}
where $\mathbf{w}_{r,k}$, $\mathbf{w}_{t,k}$ represent the $k$-th column vector in the matrices $\mathbf{W_{r}}$, $\mathbf{W_{t}}$ and the second term in (\ref{equ_5}) and (\ref{equ_6}) is the interference term under the cochannel.

The SINR of the received signal of the $k$-th user on the reflective and transmissive sides are represented respectively as
 \begin{equation}
 \begin{split}
\gamma_{r,k} = \frac{|\mathbf{H}_{RU}^T \mathbf{\Theta}_{R} \mathbf{H}_{BR} \mathbf{w}_{r,k}|^2}{\sum_{n=1, n \neq k}^{K} |\mathbf{H}_{RU}^T \mathbf{\Theta}_{R} \mathbf{H}_{BR} \mathbf{w}_{r,n}|^2 + \sigma^2},
\end{split}
\label{equ_7}
\end{equation}
 \begin{equation}
 \begin{split}
\gamma_{t,k} = \frac{|\mathbf{H}_{TU}^T \mathbf{\Theta}_{T} \mathbf{H}_{BR} \mathbf{w}_{t,k}|^2}{\sum_{n=1, n \neq k}^{K} |\mathbf{H}_{TU}^T \mathbf{\Theta}_{T} \mathbf{H}_{BR} \mathbf{w}_{t,n}|^2 + \sigma^2}.
\end{split}
\label{equ_8}
\end{equation}

This paper focuses on the sum of the users' downlink transmission rates as an indicator to evaluate the quality of the system. The $k$-th user data transfer rate on the reflective and transmissive sides is expressed, respectively, as
 \begin{equation}
\label{equ_9}
R_{r,k} = \log_2 (1 + \gamma_{r,k}),
\end{equation}

 \begin{equation}
\label{equ_10}
R_{t,k} = \log_2 (1 + \gamma_{t,k}),
\end{equation}
the sum of all users' transmission rates is expressed as
 \begin{equation}
\label{equ_11}
R = \sum_{k}^{K} (R_{r,k}+R_{t,k}).
\end{equation}

\subsection{Problem Formulation}
In this paper, we assume the CSI is known and consider optimizing the beamforming matrix $\mathbf{W}$, the phase shift matrices $\mathbf{\Theta_{R}}$, $\mathbf{\Theta_{T}}$ and the UAV's horizontal coordinates $\mathbf{C}_{R}$ to improve system performance. Under this framework, the problem is formulated as
\begin{align*}
\text{($P$1)} \max_{\mathbf{W}, \mathbf{\Theta_{R}},\mathbf{\Theta_{T}},\mathbf{C}_{R} } & \;\;\;R(\mathbf{W}, \mathbf{\Theta_{R}}, \mathbf{\Theta_{T}},\mathbf{C}_{R})  \\
\text{s.t.} & \quad tr\{\mathbf{W}\mathbf{W}^{H}\} \leq P_t \tag{12a} \label{equ_12a} \\&
\quad |\phi_{r,n}| = 1 \quad \forall n = 1, 2, \ldots, N \tag{12b} \label{equ_12b} \\&
\quad |\phi_{t,n}| = 1 \quad \forall n = 1, 2, \ldots, N \tag{12c} \label{equ_12c} \\&
\quad \beta_r^i, \beta_t^i \in [0,1] \tag{12d} \label{equ_12d} \\&
\quad \beta_r^i + \beta_t^i = 1 \tag{12e} \label{equ_12e} \\&
\quad x_{\min} < x_B < x_{\max} \tag{12f} \label{equ_12f} \\&
\quad y_{\min} < y_B < y_{\max} \tag{12g} \label{equ_12g}
\end{align*}
where $P_t$ represents the maximum output power of the BS, and (\ref{equ_12a}) indicates that the limited transmission power cannot exceed $P_t$. (\ref{equ_12b}), (\ref{equ_12c}) indicate that the amplitude of the RIS elements is $1$ to ensure that the RIS passively reflects the signals without amplifying the received signal. (\ref{equ_12d}), (\ref{equ_12e}) represent the range of the energy distribution ratio between reflection and transmission of STAR-RIS, with their sum equal to $1$. (\ref{equ_12f}) and (\ref{equ_12g}) represent the horizontal coordinate limits of the UAV. It can be seen that the problem is a non-convex problem due to the non-convexity of the objective function as well as the unit module constraint. Given the rapid development and effective application of ML algorithms in recent years, this paper uses a DRL based algorithm to find the optimal $\mathbf{W}$, $\mathbf{\Theta_{R}}$, $\mathbf{\Theta_{T}}$ and $\mathbf{C}_{R}$ in order to solve the optimization problem.

\section{SOLUTION BASED ON DRL}
In this section, we first review the principles of DRL, and then provide a detailed explanation of the CA-DDPG framework and its fundamental principles. Finally, we describe the sum rate optimization process of the RIS-assisted communication system under the CA-DDPG framework.
\subsection{Overview of DRL}
RL constitutes a crucial component of ML, aiming to identify the optimal policy for making decisions through environmental interaction. The RL framework primarily involves two parts: the agent and the environment. The agent continuously observes and analyzes the environment to determine the optimal strategy that maximizes cumulative future rewards. Conceptually, this learning paradigm aligns with the Markov decision process (MDP) framework, which models sequential decision-making in stochastic environments. The fundamental elements of RL include: states, actions, rewards, policies, and the state-action value function.

\subsubsection{State}
 $S=[s_1,s_2,...,s_t,s_{t+1},...]$ indicates the state space. State $s_t \in S$ indicates some or all cases perceived by the agent at time step $t$, and $s_{t+1}$ indicates the next state of $s_{t}$.
 \subsubsection{Action}
$A=[a_1,a_2,...,a_t,a_{t+1},...]$ represents the action space. Action $a_t \in A$ represents the measures taken by following the strategy at time instant $t$, and the state $s_{t}$ will transit to the next state $s_{t+1}$.
\subsubsection{Reward}
reward $r_t$ indicates the immediate return after the state $s_{t}$ takes the action $a_t$. It is a metric of performance that indicates the degree of effect the state $s_{t}$ takes action $a_t$.
\subsubsection{Strategy}
the probability set for the agent to take an action in a certain state is policy $\pi$, and meets $\sum_{a_t \in A} \pi(s_t, a_t) = 1$ , where $\pi(s_t, a_t)$ represents the probability of taking action $a_t$ in the state $s_{t}$.
\subsubsection{State-action value function}
agents may fall into a ``myopic dilemma'' when focusing solely on immediate rewards.  Therefore, in addition to focusing on immediate rewards, we should go a step further and consider future rewards. $Q_\pi(s_t,a_t)$ is defined as the state-action value function, which represents the expectation of immediate rewards and future cumulative rewards received given the state $s_{t}$ and action $a_t$. The $Q$ function of executing policy $\pi$ can be expressed as
 \begin{equation}
\label{equ_13}
Q_{\pi}(s_t, a_t) = \mathbb{E}_{\pi}[R_t | s = s_t, a = a_t],\tag{13}
\end{equation}
where $R_t = \sum_{\tau=0}^{\infty} \lambda ^{\tau} r_{t+\tau+1}$
 represents the sum of future discount rewards, and $\lambda \in (0,1]$ is the discount rate.

 The $Q$ function satisfies the Bellmann equation as follows
 \begin{align*}
Q_{\pi}(s_t, a_t) &= \mathbb{E}_{\pi}[r_{t+1} | s = s_t, a = a_t] +  \lambda \sum_{s_{t+1} \in S} P_{(s_t \to s_{t+1})}^{a} \\& \times\left( \sum_{a_{t+1} \in A} \pi(s_{t+1}, a_{t+1}) Q_{\pi}(s_{t+1}, a_{t+1}) \right), \tag{14}
\end{align*}
where $P_{(s_t \to s_{t+1})}^{a} = \mathbb{P}(s' = s_{t+1} | s = s_t, a = a_t)$
 represents the state transition probability of taking the action $a_{t}$ from state $s$ to state $s_{t+1}$ and state $s'$ represents the next state of $s$.

Using the method of Q-learning to find the optimal policy , the optimal $Q$ function can be rewritten as
\begin{align*}
Q_{\pi}^*(s_t, a_t) = & r_{t+1}(s = s_t, a = a_t, \pi = \pi^*) + \\
& \lambda \sum_{s_{t+1} \in S} P_{(s_t \to s_{t+1})}^a \max_{a_{t+1} \in A} Q_{\pi}^*(s_{t+1}, a_{t+1}). \tag{15}
\label{equ_10}
\end{align*}

The Bellman equataion (\ref{equ_10}) can be solved recursively, and the recursive update rule of $Q$ function is as follows
\begin{align*}
Q_{\pi}^{*}(s_{t}, a_{t}) \leftarrow & (1 - \beta) Q_{\pi}^{*}(s_{t}, a_{t}) + \\
& \beta (r_{t+1} + \lambda \max_{a_{t} \in A} Q_{\pi}(s_{t+1}, a_{t})),   \tag{16}
\end{align*}
where $\beta$ represents the learning rate of the $Q$ function.

Q-learning can efficiently accomplish tasks when the state space and action space are small or finite. However, when the state and action spaces become large or even infinite, the $Q$ table that records all $Q$ values in Q-learning would require an enormous storage capacity, which is clearly impractical. To address this issue, the deep Q network (DQN) was developed, which replaces the $Q$ table with a deep neural network (DNN). The DNN estimates the system model as a combination of multiple nonlinear functions through policy functions, action functions, and state-value functions, where both actions and $Q$ function are represented by DNNs rather than explicit mathematical models. The DNN takes the observed state of the environment as input and produces actions as output. Another DNN estimates the $Q$ values of the actions. For DRL, the neural network is trained with the parameter $\mu$ to evaluate the $Q$ value, so the $Q$ function of the agent can be represented by the parameter $\mu$ as follows
\begin{align*}
Q_{\pi}(s_t, a_t; \mu) = \mathbb{E}_{\pi}[R_t | S = s_t, A = a_t],
\tag{17}   
\end{align*}
where $\mu$ represents the bias and weighting parameters of the neural network in DQN, and the optimal $Q$ function is obtained by using the stochastic optimization algorithm, rather than by directly updating the $Q$ function. The parameter $\mu$ update rule is as follows
\begin{equation}
\label{deqn_ex1a}
\mu_{t+1} = \mu_t + \rho \nabla_\mu \ell(\mu),\tag{18}
\end{equation}
where $\rho$ represents the learning rate of updating $\mu$, and $\nabla_\mu$ represents the gradient of the loss function $\ell(\mu)$.

The loss function is defined as the difference between the predicted value and the actual target value of the neural network output. The neural network in the DQN method takes the state as input and then estimates the maximum action with the $Q$ value function as the output action. After each round, the parameter $\mu$ of the neural network is updated to obtain a more accurate $Q$ value estimate. Since the actual target value is unknown in RL, it is necessary to define two neural networks with identical structures: a training network and a target network, parameterized by $\mu_{(train)}$ and $\mu_{(target)}$, respectively. The difference between the predicted values and the actual target values is expressed by the loss function as
\begin{align*}
\ell(\mu) = \left( y - Q_{\pi}(\mu_{(train)} | s_t, a_t) \right)^2 
\tag{19}   
\end{align*}

The target neural network is synchronized with the training neural network update at a certain frequency, and the actual target value $y$ is defined as
\begin{align*}
y = r_{t+1}(s_t, a_t) + \lambda \max_{a_t' \in A} Q_{\pi}(\mu_{\text{(target)}} | s_{t+1}, a_t'). 
\tag{20}   
\end{align*}

Compared to DQN, the DDPG algorithm, based on the actor-critic framework can handle continuous action spaces and employs a deterministic policy that maps a given state to a specific action. The DDPG algorithm consists of four DNNs: the training actor network, the target actor network, the training critic network and the target critic network. The actor network generates actions based on the current state, while the critic network is responsible for evaluating based on the performance of states and actions. The action is approximated using the participant network, thus eliminating the need for a next non-convex optimization step to find the maximum $Q$ value function. The update rules for the training critic network are as follows
\begin{align*}
\mu_{t+1}^c = \mu_t^c - \xi^c \nabla_{\mu_{(train)}^c} \ell(\mu_{(train)}^c),
\tag{21}   
\end{align*}

\begin{align*}
\ell(\mu_{\text{(train)}}^{c}) = \left( r_{t} + \lambda Q_{\pi}(\mu_{\text{(target)}}^{c} | s_{t+1}, a_{t}^{'}) - Q_{\pi}(\mu_{\text{(train)}}^{c} | s_{t}, a_{t}) \right)^{2},
\tag{22}   
\end{align*}
where $\xi^c$ is the learning rate of the training critic network. $\nabla_{\mu_{(train)}^c} \ell(\mu_{(train)}^c)$ represents the gradient of the training critic network $\mu_{(train)}^c$. $a_{t}^{'}$ is the action output of the target actor network. The parameters of the target network are updated after a certain number of episodes to synchronize with the parameters of the training network, and the update rule for the training actor network is as follows
\begin{align*}
\label{equ_18}
\mu_{t+1}^{a} = \mu_{t}^{a} - \xi^{a} \nabla_{a} Q_{\pi}(\mu_{\text{(target)}}^{c} | s_{t}, a_{t}) \nabla_{\mu_{\text{(train)}}^{a}} \pi(\mu_{\text{(train)}}^{a} | s_{t}),
\tag{23}   
\end{align*}
where $\xi^{a}$ is the learning rate of the actor network, $\nabla_{a} Q_{\pi}(\mu_{\text{(target)}}^{c} | s_{t}, a_{t})$ and $\nabla_{\mu_{\text{(train)}}^{a}} \pi(\mu_{\text{(train)}}^{a} | s_{t})$ represent the gradients of the target critic network and the training actor network with respect to their parameters $\mu_{\text{(target)}}^{c}$ and $\mu_{\text{(train)}}^{c}$, respectively. As can be seen from (\ref{equ_18}), the parameter update strategy is influenced by the gradient of the action to ensure that the next action is selected according to the direction of the optimal action strategy, thus optimizing the $Q$ function.

The update rules for the target actor and critic networks are as follows
\begin{align}
\label{equ_19}
\mu^{a}_{(\text{target})} &\leftarrow (1 - \tau)\mu^{a}_{(\text{target})} + \tau\mu^{a}_{(\text{train})},
\tag{24} 
\end{align}
\begin{align}
\label{equ_20}
\mu^{c}_{(\text{target})} &\leftarrow (1 - \tau)\mu^{c}_{(\text{target})} + \tau\mu^{c}_{(\text{train})}, 
\tag{25} 
\end{align}
where $\tau$ is the soft update coefficient of the target actor and critic networks.

\subsection{CA-DDPG Framework}
\begin{figure}[t]
\centering
\includegraphics[width=3.5in]{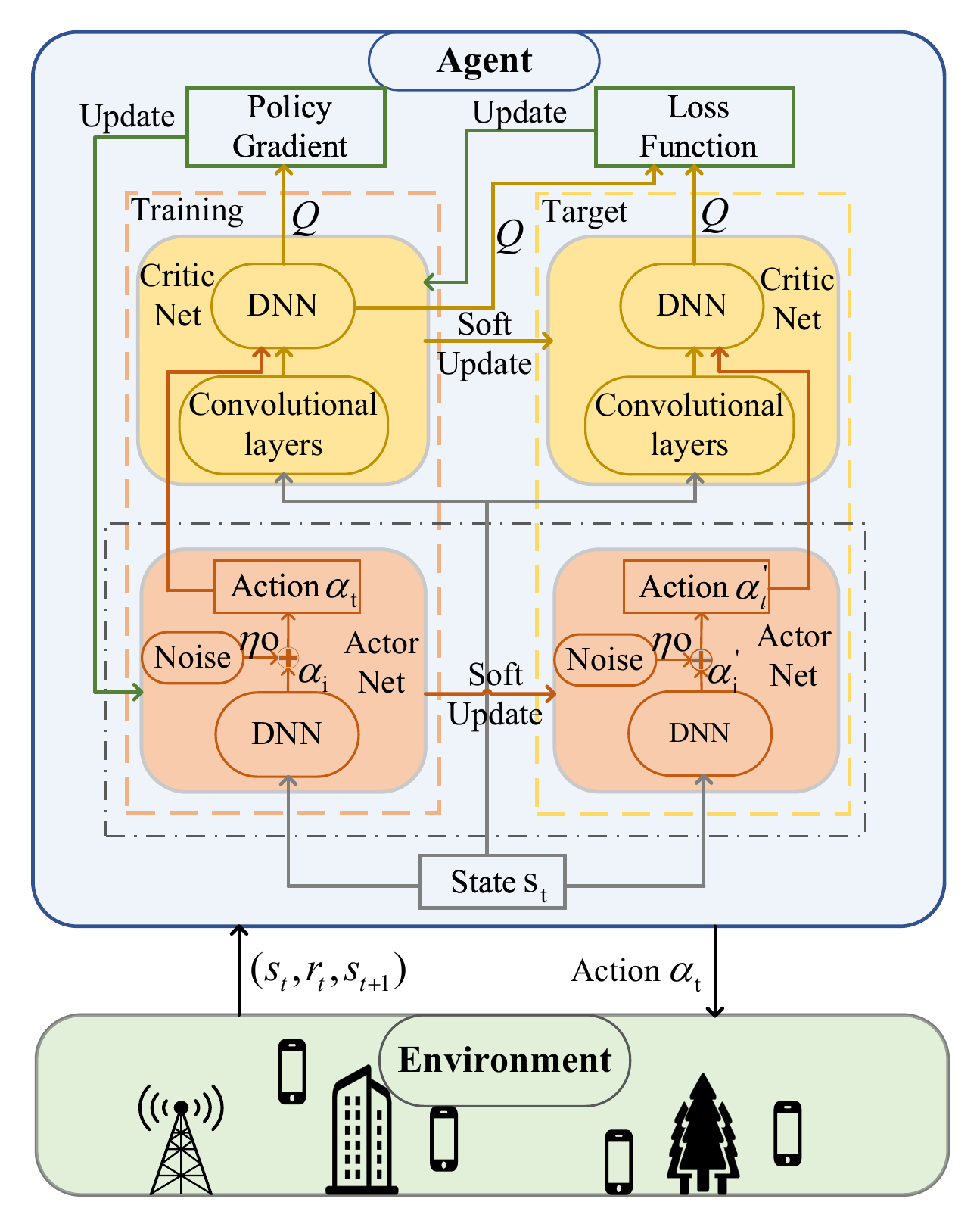}
\caption{CA-DDPG framework.}
\label{fig.2}
\end{figure}

The Q-learning method is straightforward and easy to implement, but it is not an optimal choice for complex, high-dimensional problems. While DQN can address the traditional curse of dimensionality in high-dimensional spaces and demonstrates strong effectiveness in handling discrete state problems, its performance degrades significantly in continuous action domains. The DDPG algorithm employs the actor-critic architecture, which can effectively handle continuous control tasks and enhance the efficiency of policy updates. Building on DDPG, our proposed CA-DDPG algorithm has been modified in the following aspects. To address DDPG’s limitations in action exploration efficiency, we introduce a stochastic noise factor into the actor network to facilitate a more comprehensive exploration of the action space. As the number of actions increases, the original critic network's evaluation performance becomes insufficient, and its ability to distinguish between good and bad actions is suboptimal. Therefore, we design an improved critic network to enhance the evaluation capability. This architecture significantly boosts the accuracy of the estimation of $Q$ value by precisely differentiating between better and suboptimal state-action pairs, ultimately enhancing overall performance of the system. The detailed framework of the CA-DDPG algorithm is illustrated in Fig.\ref{fig.2}.

\subsubsection{State space}
the state $s_t$ at time step $t$ consists of the action $a^{(t-1)}$ from the previous time step $(t-1)$, transmit power $T_p$, receiving power $R_p$, the channel state matrices $\mathbf{H}_{BR}$ and $\mathbf{H}_{RU}$. The transmitting power $T_p$ of the $k$-th user and receiving power $R_p$ are expressed as
\begin{align}
T_p &= ||W_k||^2 = |W_k^{\mathrm{H}} W_k|^2
\tag{26} 
\end{align}
\begin{align}
R_p &= |H_{RU}^T \Theta H_{BR} W|^2.
\tag{27} 
\end{align}

Since the neural network can only accept real numbers as inputs, when constructing the state $s$, any complex numbers must be split into their real and imaginary parts before being used as inputs. Given the transmission symbol with unit variance, the transmission power of the $k$-th user is given by $\left\| \mathbf{W}_k \right\|^2 = \left| \text{Re} \{ \mathbf{W}_k^{\text{H}} \mathbf{W}_k \} \right|^2 + \left| \text{Im} \{ \mathbf{W}_k^{\text{H}} \mathbf{W}_k \} \right|^2
$. Let $\mathbf{P}_{W}(n) = \mathbf{H}_{RU}^{T} \mathbf{\Theta H}_{BR} \mathbf{W}(n)$, the receiving power of the $k$-th user is defined as $R_p(n) = | \text{Re} \{ \mathbf{P}_W(n) \} |^2 + | \text{Im} \{ \mathbf{P}_W(n) \} |^2$. The first term of the transmission power and receiving power is the real part, and the second term is the imaginary part. These parts serve as independent input ports for the actor network and the critic network. Therefore, the state at time step $t$ is defined as
\begin{align}
s_t = [a^{(t-1)}, T_p, R_p, \mathbf{H}_{BR}, \mathbf{H}_{RU}, \mathbf{H}_{TU}].
\tag{28} 
\end{align}

\subsubsection{Action space}
the action $a_t$ at time step $t$ is constructed from the transmission beamforming matrix $\mathbf{W}$ and the phase shift matrix $\mathbf{\Theta}$. Similarly, $\mathbf{W}$ and $\mathbf{\Theta}$ are decomposed as $\mathbf{W}=Re(\mathbf{W})+Im(\mathbf{W})$ and $\mathbf{\Theta}=Re(\mathbf{\Theta})+Im(\mathbf{\Theta})$, respectively. The action $a_t$ of the time step $t$ is expressed as
\begin{align}
a_t = [\mathbf{W}^t, \mathbf{\Theta_{R}}^t, \mathbf{\Theta_{T}}^t, \mathbf{C_{R}}^t].
\tag{29} 
\end{align}

\subsubsection{Reward}
when the agent selects an action at time step $t$, the reward $r_t$ is used to evaluate the quality of the strategy. In this paper, our goal is to find the optimal system sum rate $R$, so the reward function is represented by the system sum rate $R$ as follows
\begin{align}
r_t = R^t.
\tag{30} 
\end{align}

\subsection{Framework of Convolutional Layer}
The CNN represents a class of feedforward neural networks inspired by biological visual perception mechanisms. The core components of CNN include convolutional layers, activation functions, pooling layers, fully connected layers, and normalization layers.

\begin{figure}[t]
\centering
\includegraphics[width=3.5in]{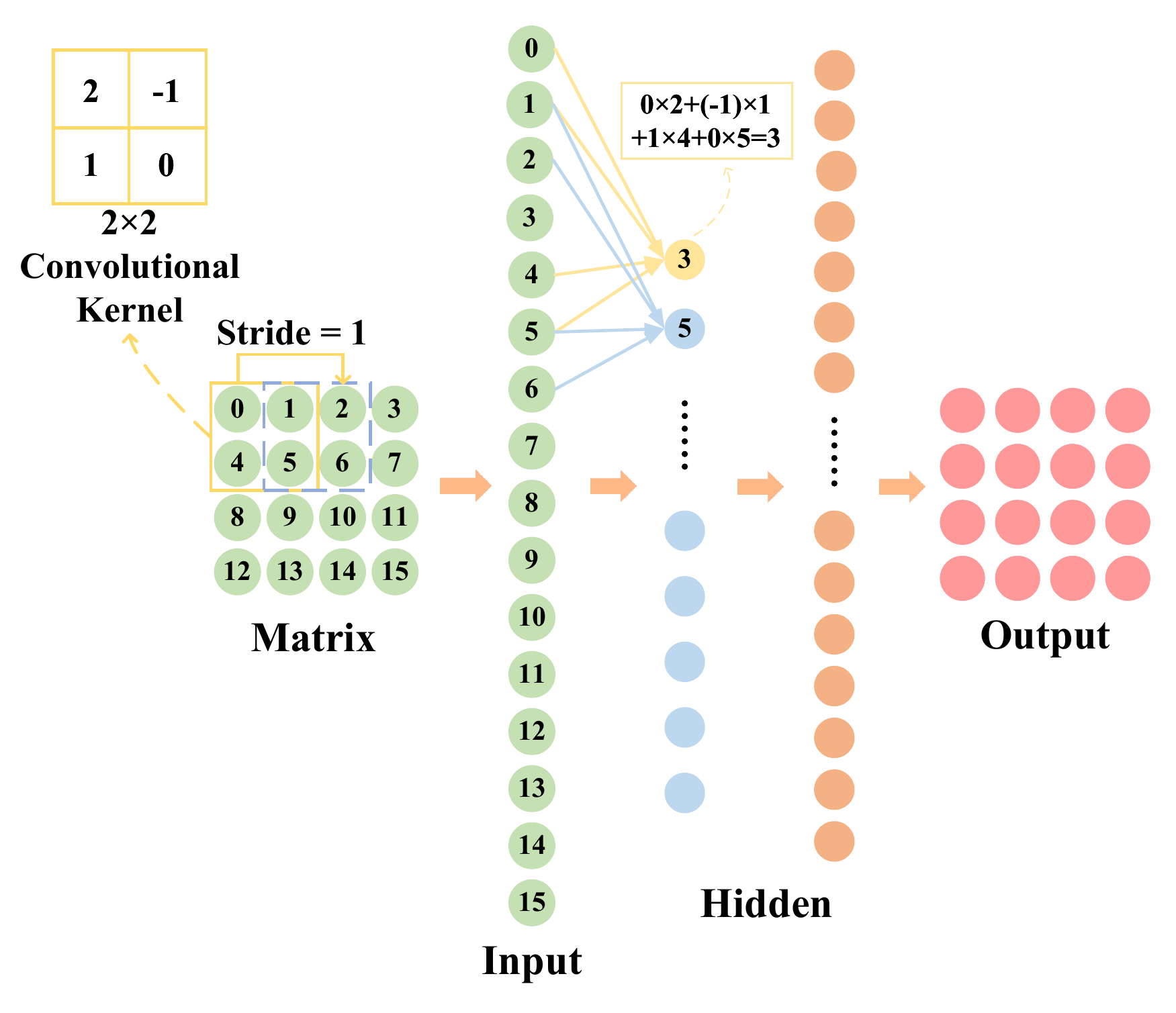}
\caption{Basic framework of convolution.}
\label{fig.3}
\end{figure}

This paper mainly applies convolutional layers to enhance the critic network. Specifically, convolutional layers comprise key components such as kernels, kernel sizes, strides, and padding configurations. The basic framework of convolution is shown in Fig.\ref{fig.3}. The mathematical foundation of CNN lies in convolution operations, where multiple learnable filters are applied to input data to automatically extract hierarchical features. These filters slide across the spatial dimensions of the input, computing dot products within local receptive fields to construct hierarchical feature representations. As the depth of the network increases, the model progressively captures features ranging from simple patterns to complex semantic representations, enabling tasks such as classification and recognition. Furthermore, pooling operations reduce the spatial dimensions of feature maps while enhancing invariance to input transformations. The entire training process is driven by backpropagation, where filter weights are optimized through loss function minimization to reduce prediction errors.
\subsubsection{Convolutional Kernel}
The convolutional kernel is a core component of the convolutional layer. Essentially, a convolutional kernel is a small weight matrix used to extract local features from the input data. Its operating principle is similar to that of a ``sliding window filter'', which progressively scans the input data through convolution operations and to generate a feature map. The convolutional kernel is typically a square matrix, with each corresponding position storing a learnable weight coefficient. The kernel size is usually $n \times n$, which determines the size of the receptive field. A convolutional layer can contain multiple kernels simultaneously, and each kernel generates an independent feature map. Additionally, the weight coefficients of the convolutional kernel are automatically optimized through backpropagation, eliminating the need for manual design.
\subsubsection{Stride}
The stride determines the distance that the convolutional kernel moves during each step of the convolution process. The kernel slides over the input according to the stride, performing dot product and summation operations region by region to compute a weighted combination of local features.
\subsubsection{Padding}
Padding refers to adding extra values around the edges of the input data to adjust its dimensions while preserving edge information.

The dimensional transformations in a convolutional layer follow the principle given by
\begin{align}
J' = \frac{(J + 2p - k_s)}{s_d} + 1, 
\tag{31} 
\end{align}
where $J$ represents the input matrix size, $J^{'}$ represents the output matrix size, $p$ represents the number of padding, $k_s$ represents the convolution kernel size, and $s_d$ represents the stride.

The convolutional layer can automatically extract key features through local receptive fields and weight sharing. The discriminative features extracted by the convolutional layer enable the critic to discern subtle state differences, thereby providing a more reliable policy gradient. Meanwhile, the translation invariance of the convolution makes the network more robust to input noise, enabling the policy to remain stable in complex environments. By enhancing the critic's state feature extraction capabilities through convolutional layer, the network can more accurately capture critical information within states. Combined with the global integration of fully connected layers and action fusion, this ultimately achieves precise $Q$ value estimation.

\subsection{Algorithm Description}
Our proposed algorithm consists of four networks: actor network $\mu^a_{\text{(train)}}$, actor target network $\mu^a_{\text{(target)}}$, critic network $\mu^c_{\text{(train)}}$, and critic target network $\mu^c_{\text{(target)}}$. The parameters of the actor target network and critic target network are obtained by copying the actor and critic networks.

At the beginning of the algorithm, the experience replay buffer $B$, actor network $\mu_{\text{(train)}}^{a}$, critic network $\mu_{\text{(train)}}^{c}$, precoding matrix $\mathbf{W}$ and phase shift matrix $\mathbf{\Theta}$ should be initialized first. In this paper, we simply initialize the precoding matrix $\mathbf{W}$ and the phase shift matrix $\mathbf{\Theta}$ with Gaussian random matrices that satisfies the constraints. Next, the initial state $s_t$ is input to the actor network to generate an action $a_i$. This action is then perturbed by random factors to produce the final selected action $a_t$. The specific rule for generating action $a_t$ is as follows
\begin{equation}
a_t = a_i + \eta o,
\tag{32}
\end{equation}
where $\eta$ represents the decreasing coefficient of the random factor, which decreases with the number of time steps. $o$ represents a random factor that follows a normal distribution and has the same format as the action $a_t$. After performing the action, we obtain the next state $s_{t+1}$ and reward $r_t$, and then we save the resulting data as tuples $(s_t, a_t, r_t, s_{t+1})$ into the experience replay buffer $B$. When calculating the $Q$ value, a small batch of tuples $(s_t, a_t, r_t, s_{t+1})$ is extracted from the experience replay buffer $B$ to train the critic network to receive the state $s_{t+1}$ and action $a_{t+1}$ as input. The state is first processed by the convolutional layers, after which the output is combined with the action and both are fed into the linear neural network to compute the target $Q$ value. The final target $Q$ value is obtained by multiplying the target $Q$ value with the discount coefficient and adding the reward $r_t$.

At the same time, the current $Q$ value is obtained with state $s_t$ and action $a_t$ as input. The loss of the training critic network is calculated using the mean square loss function, and optimize the parameters of the training critic network using backpropagation loss. In contrast, the loss of the training actor network is obtained by the state $s_t$ and action $a_t$ input to the training critic network to optimize the parameters of the training actor network. After a certain number of steps, the parameters of the target actor and critic networks are softly updated with the training network parameters. The update rules as consistent as (\ref{equ_19}) and (\ref{equ_20}). Algorithm \ref{algorithm1} outlines the steps of the CA-DDPG framework.

\begin{algorithm}[t]
	\caption{CA-DDPG Framework}
	\label{algorithm1}
	\begin{algorithmic}[1]
		\Require 
            Channel state information $(\mathbf{H}_{BR}, \mathbf{H}_{RU}, \mathbf{H}_{TU})$ of all \Statex \hspace*{-7.2mm} the users.    

\Statex \hspace*{-6mm}\textbf{Initialization:} 
beamforming matrix $\mathbf{W}$, phase shift matrix \Statex \hspace*{-7.2mm} $\mathbf{\Theta_{R}}$ and $\mathbf{\Theta_{T}}$, UAV's horizontal coordinate  $\mathbf{C_{R}}$, \Statex \hspace*{-7.2mm} experience replay buffer $B$, training actor network para-\Statex \hspace*{-7.2mm} meter  $\mu^{a}_{\text{(train)}}$, training critic network parameter $\mu^{c}_{\text{(train)}}$, target \Statex \hspace*{-7.2mm} actor network parameter $\mu^{a}_{\text{(target)}}$, target critic network para- \Statex \hspace*{-7.2mm} meter $\mu^{c}_{\text{(target)}}$. 

		\Ensure 
            optimal action $a_t$, sum rate $R$ and $Q$ value function.
            
		\For{each episode}
		\State Get channel information $\mathbf{H}_{BR}$, $\mathbf{H}_{RU}$, $\mathbf{H}_{TU}$
		\State Obtain the initial state $s_0 \in S, S \leftarrow s_0$.
		\For{{each time step}}
		\State Observe the initial state $s_t$.
            \State Obtain action $a_i$.
            \State Add noise random factor $o$ to $a_i$.
            \State Execute the action $a_t=a_i+\eta o$.
            \State Observe next state $s_{t+1}$ and reward $r_{t}$.
            \State Store the experience $(s_t, a_t, r_t, s_{t+1})$ into $B$.
            \State Sample a minibatch randomly form $B$.
            \State Input the state $s_t$ and convolve it through convolutional layers.
            \State Obtain the $Q$ value function from the critic network.
            \State Update the parameters of the critic network by minimizing the critic's loss via equation (22).
            \State Update the parameters of actor network with sampled policy gradients by equation (23).
            \State Every $J$ steps, soft update the parameters of the target networks by equation (24) and (25).
            \State Update the state $s_{t+1}$.
            \EndFor
            \EndFor

	\end{algorithmic}
\end{algorithm}

\subsection{Complexity Analysis}
In general, the complexity of the DDPG algorithm is determined by the specification of the DNN. For the entire network, if we assume that the number of neurons in each layer is \( u \), the total number of parameters is approximately \( O(Lu^2) \), where \( L \) is the number of neural network layers. The state space dimension of DDPG algorithm is \( D_s = 3K + 2NM + 4NK + 2MK + 4N +2\). Therefore, the complexity of the DNN is about \( O(LD_s^2) \), and the total complexity of the training actor-critic network and the target actor-critic network is \( O(4LD_s^2) \), so the complexity of the DDPG algorithm is \( O(4LD_s^2) \). The TD3 algorithm has six neural networks, including two actor networks and four critic networks, so the complexity of the TD3 algorithm is \( O(6LD_s^2) \). The proposed CA-DDPG algorithm incorporates convolutional layers into the critic network, with the same number of convolutional layers as DNN and the number of channels set to \( D_s\). In this paper, we assume that the number of convolutional layers is \( L' \) and \( L' \) less than $L$. Therefore, the complexity of the CA-DDPG algorithm is \( O(4LD_s^2+2L'D_s^2) \).

\section{NUMERICAL EVALUATION AND DISCUSSION}
In this section, we perform a simulation of the STAR-RIS-UAV-assisted wireless communication system. To evaluate the performance of the proposed algorithms, we compare them with the DDPG algorithm and some new DRL algorithms. We assume that the system consists of a BS with $M$ antenna, $K$ single antenna users on the reflecting side and the same $K$ single antenna users on the transmitting side, a UAV located between the base station and the users, and equipped with STAR-RIS elements of $N$ RIS units. The BS’s coordinates are set to $[0,0]$, the coordinates of all reflective users are $[80,0]$, the coordinates of the transmissive users are $[80,80]$, the initial coordinates of the UAV are $[40,20]$, with a fixed height of 30 meters. Moreover, we assume that the channel state matrices $\mathbf{H}_{BR}$, $\mathbf{H}_{RU}$ and $\mathbf{H}_{TU}$ follow the rayleigh distribution, and the transmission power does not exceed $P_t$. The input of the actor network is the state, and the size is the dimension of the state space. The actor network performs non-linear activation using the tanh activation function, and the output is an action vector. The critic network has two parts, the convolutional layer and the linear layer. the convolutional layer is activated using the ReLU function, and the linear layer is activated using the tanh activation function. Moreover, the parameters of the actor and critic networks are updated using the Adam optimizer. The parameter list for the CA-DDPG algorithm is detailed in Table \ref{table1}. In addition, this paper also records the runtime of various algorithms to demonstrate their execution speed. Table \ref{table2} provides a detailed explanation of the runtime.

\begin{table}[t]
\begin{center}
\caption{\\SIMULATION PARAMETERS} 
\small
\begin{tabular}{|m{1.2cm}|m{5cm}|m{1cm}|}
\hline
Parameter & Description & Value\\
\hline
$B$ & Experience replay buffer size & 80000\\
\hline
$b_{m}$ & Mini-batch size & 16\\
\hline
$E_{p}$ & Maximum number of steps per episode & 80000\\
\hline
$\tau$ & Soft update rate & 0.001\\
\hline
$\xi^{a}$ & Actor learning rate  & 0.001\\
\hline
$\xi^{c}$ & Critic learning rate  & 0.001\\
\hline
$d_{c}$ & Decaying rate for training critic network uptate & 0.00001\\
\hline
$d_{a}$ & Decaying rate for training actor network uptate & 0.00001\\
\hline
$\lambda$ & Discount factor  & 0.8\\
\hline
$k_{s}$ & Convolution kernel size & 1\\
\hline
$p$ & Convolutional layer's padding  & 0\\
\hline
$s_{d}$ & Convolutional layer's stride  & 1\\
\hline
$J$ & The number of steps for synchronizing updates between the training network and the target network & 1\\
\hline
\end{tabular}
\label{table1}
\end{center}
\end{table}

\begin{table}[t]
\centering
\caption{\\RUNTIME} 
\begin{tabular}{|l|l|l|}
\hline
Algorithm & Complexity & Runtime Per Episode(second) \\ \hline
CA-DDPG & \( O(4LD_s^2+2L'D_s^2) \) & 1836.2 \\ \hline
TD3 & \( O(6LD_s^2) \) & 2020 \\ \hline
DDPG & \( O(4LD_s^2) \) & 1460.8 \\ \hline
\end{tabular}
\label{table2}
\end{table}

\subsection{Algorithm Overview}
To validate the algorithm's performance, we conducted comparative experiments with several methods, including the classical DDPG algorithm and recently proposed algorithms:

\subsubsection{DDPG}
The DDPG algorithm, which effectively solves problems in continuous action spaces, is based on the actor-critic framework. It has been successfully applied to optimize beamforming and phase shift matrices for maximizing the sum rate in RIS-aided wireless communication systems.
\subsubsection{DDPG-based DTDE}
The DDPG-based DTDE framework, which is built upon DDPG, maximizes the sum rate by optimizing the system's beamforming matrix and phase shift matrix. \cite{10261216}.
\subsubsection{TD3}
The TD3 algorithm employs six neural networks, including two actor networks and four critic networks. It mitigates overestimation issues through delayed policy updates and target policy smoothing, optimizing the beamforming and phase shift matrices in RIS-assisted systems.

\begin{figure}[t]
\centering
\includegraphics[width=3.5in]{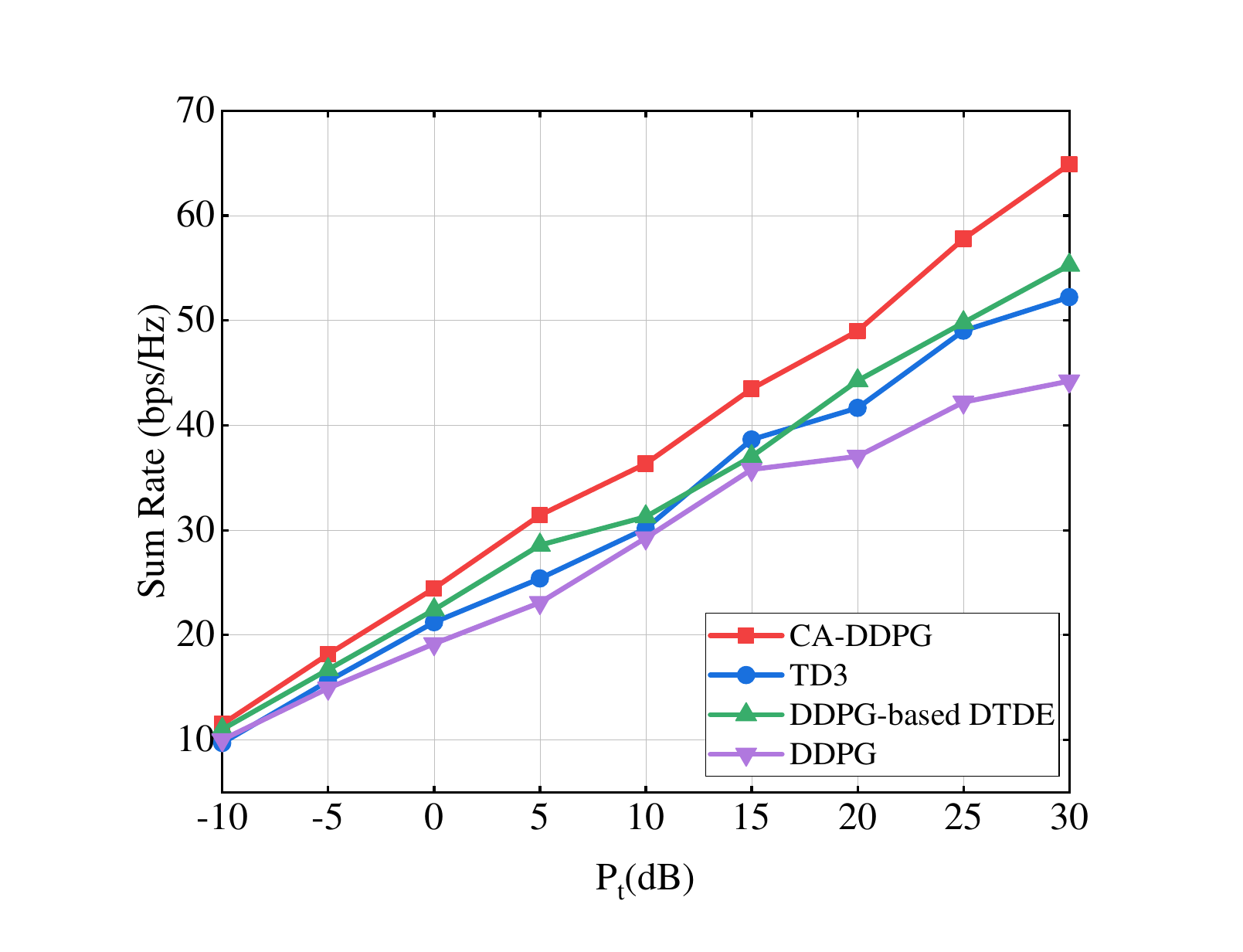}
\caption{Sum rate of four algorithms under different $P_t$.}
\label{fig.4}
\end{figure}

\subsection{Numerical Analysis}
As shown in Fig.\ref{fig.4}, we evaluate the relationship between transmit power variations and the maximum sum rate for the proposed CA-DDPG algorithm, with a comparative analysis against multiple DRL algorithms. The experimental environment is uniformly configured with $M=4$, $K=4$, and $N=16$. The results demonstrate that the sum rate consistently improves as transmit power increases. Recently proposed algorithms exhibit superior performance compared to classical DRL methods, while the CA-DDPG algorithm outperforms other algorithms under both low-power and high-power conditions.

\begin{figure}[t]
\centering
\includegraphics[width=3.5in]{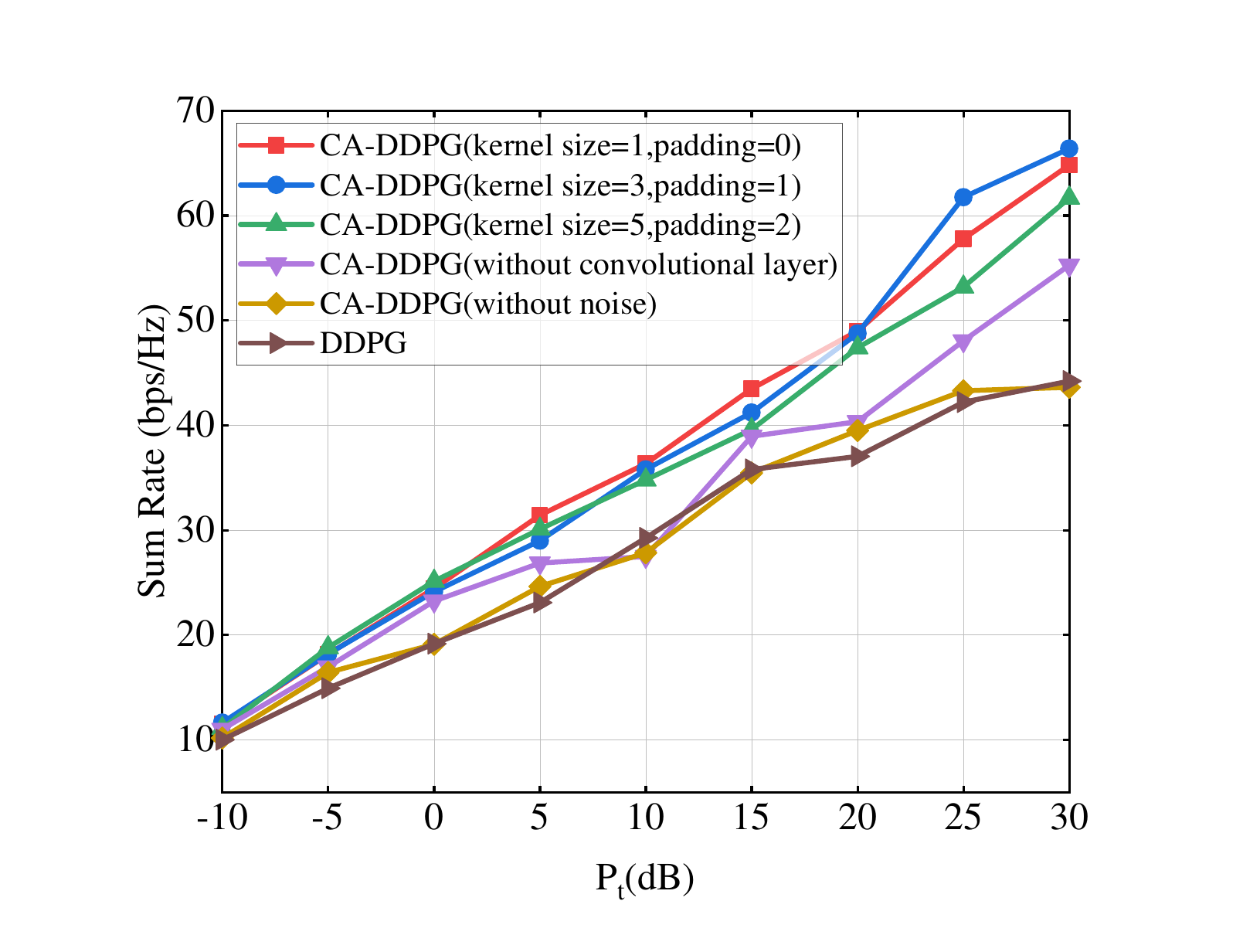}
\caption{Ablation experiment of the proposed algorithm.}
\label{fig.5}
\end{figure}

To further verify the effectiveness of the proposed algorithm, we decompose it and conduct ablation experiments. As shown in Fig.\ref{fig.5}, simply increasing the convolutional layers does not improve the sum rate, while the introduction of the noise factor effectively enhances the system's sum rate. Our proposed algorithm combines these two components and the experimental results demonstrate a significant performance improvement. In addition, we adjust the parameters of the convolutional layers. The simulation results indicate that changes in kernel size and padding have a certain impact on the experimental results, but the overall performance remains relatively consistent. It can be inferred that using appropriate parameters under different signal-to-noise ratio (SNR) conditions can yield better performance gains.

\begin{figure}[t]
\centering
\includegraphics[width=3.5in]{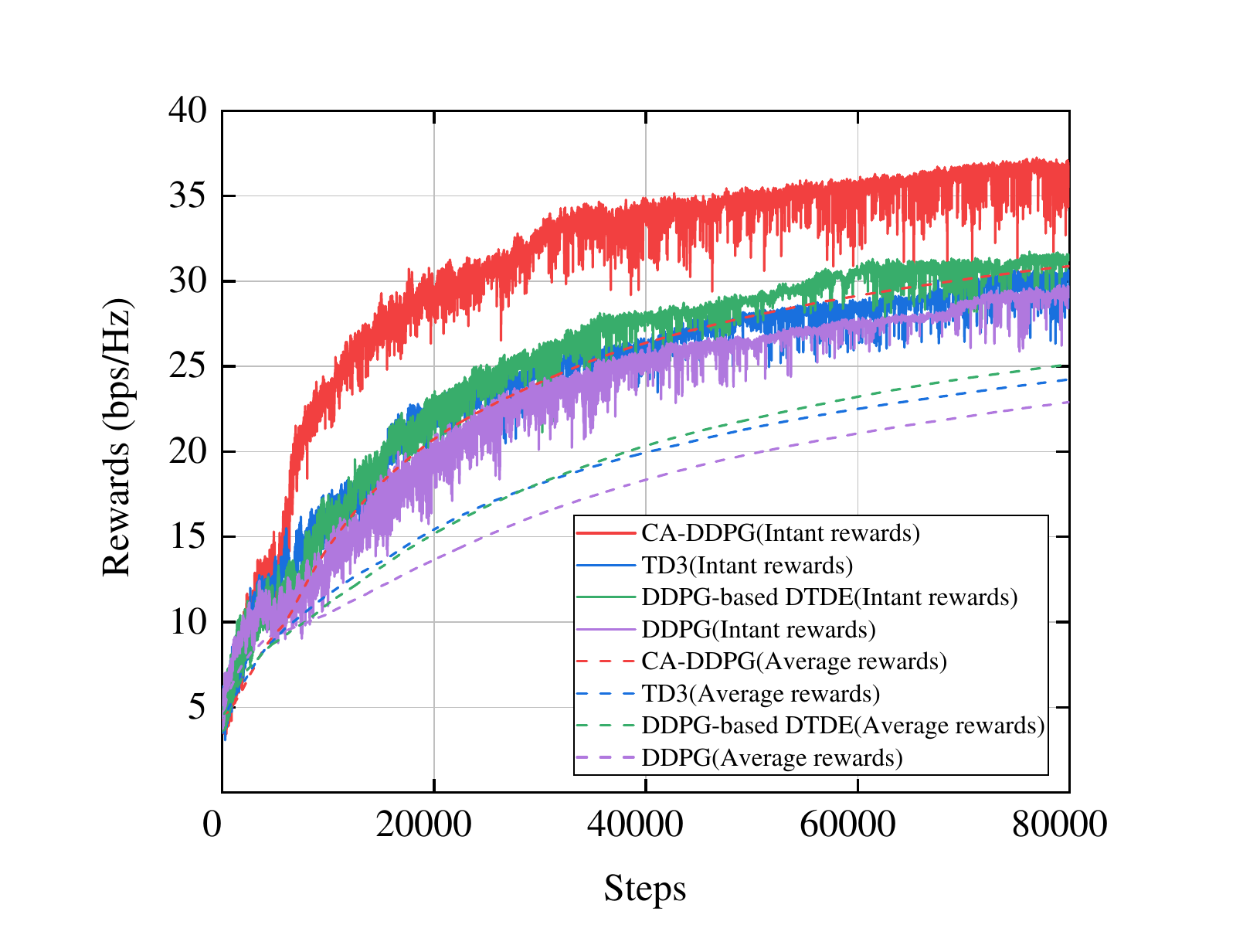}
\caption{Rewards versus time steps under different algorithms.}
\label{fig.6}
\end{figure}

To further investigate the performance of different algorithms with the same power, we focused on the results of instantaneous rewards and average rewards. The method for calculating the average reward is as follows
\begin{equation}
\label{equ33}
\text{Avg\_reward}(E_i) = \frac{\sum_{e=1}^{E_i} \text{reward}(e)}{E_i},\ E_i = 1,2,\dots, E_{p},
\tag{33}
\end{equation}
where $E_i$ represents the $i$-th step, and $E_{p}$ represents the maximum number of steps. Fig.\ref{fig.6} illustrates the relationship between time steps and instantaneous rewards for different algorithms under identical SNR conditions. As shown in the figure, at $P_t=10dB$, all algorithms exhibit increasing rewards with time steps, and the reward approaches the convergence value near $Steps=40000$ and eventually stabilizes. Notably, the CA-DDPG algorithm consistently achieves higher rewards than all baseline methods. 

\begin{figure}[t]
\centering
\includegraphics[width=3.5in]{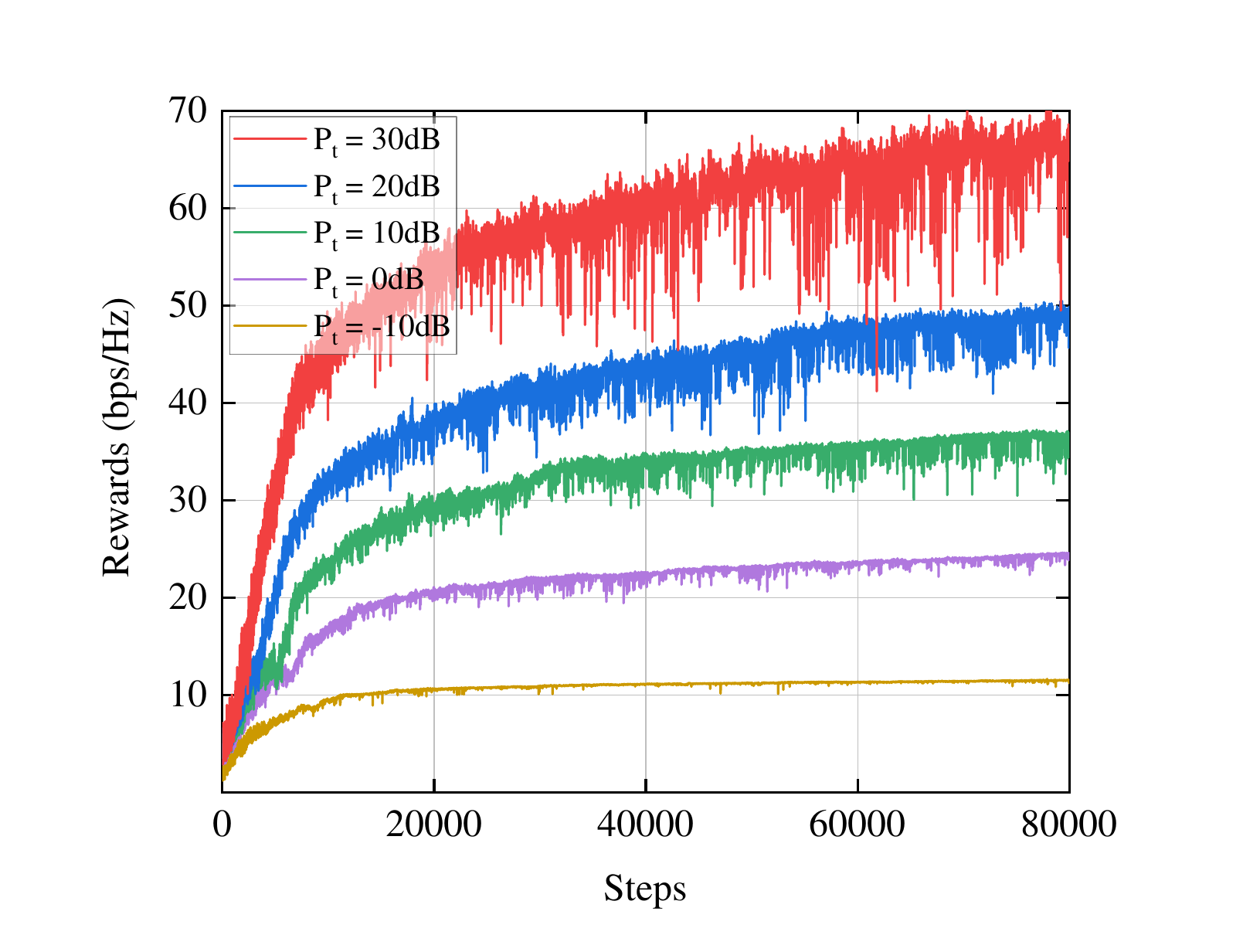}
\caption{Rewards versus time steps under different $P_t$.}
\label{fig.7}
\end{figure}

To investigate the CA-DDPG algorithm, we examined the rewards at $P_{t} = \{-10,0,10,20,30\} dB$ , as shown in Fig.\ref{fig.7}. The results reveal that the sum rate gradually increases with training steps and eventually converges. The transmit power level creates distinct stratification in sum rate performance, where higher transmit power yields greater sum rate. Moreover, the reward curve under lower transmit power exhibits smoother fluctuations, whereas the curve becomes increasingly volatile at higher power levels.

\begin{figure}[t]
\centering
\includegraphics[width=3.5in]{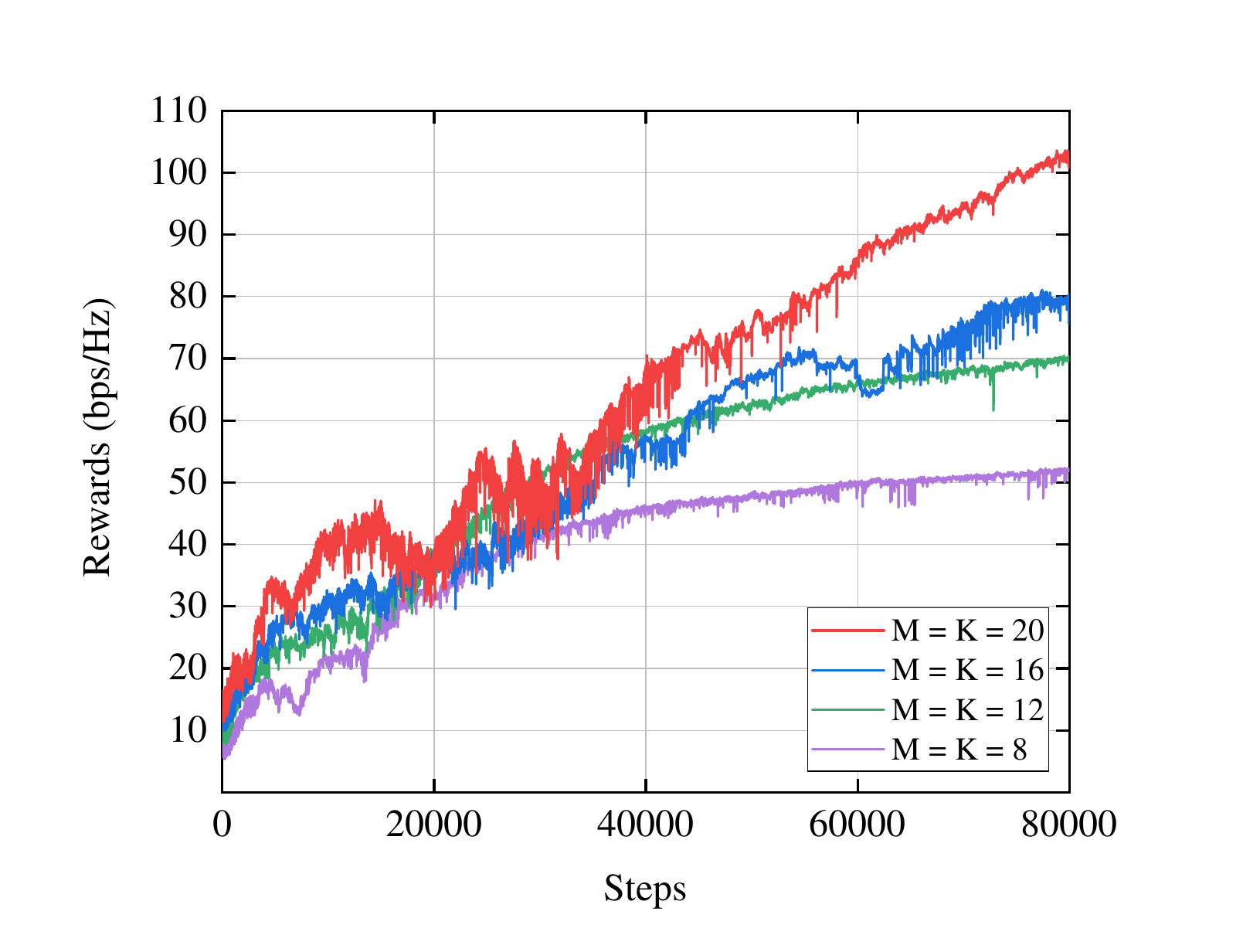}
\caption{Rewards versus time steps under different $M$ and $K$.}
\label{fig.8}
\end{figure}

Fig.\ref{fig.8} shows the relationship between the number of BS antennas and users and the sum rate. The experimental environment is set to $M=K= \{8,12,16,20\}$ and $N=16$. As shown in the figure, as the number of BS antennas and users increases, the sum rate under multi-antenna and multi-user scenarios is higher at convergence. It should be noted that convergence occurs more quickly when the number of antennas and users is smaller.

\begin{figure}[t]
\centering
\includegraphics[width=3.5in]{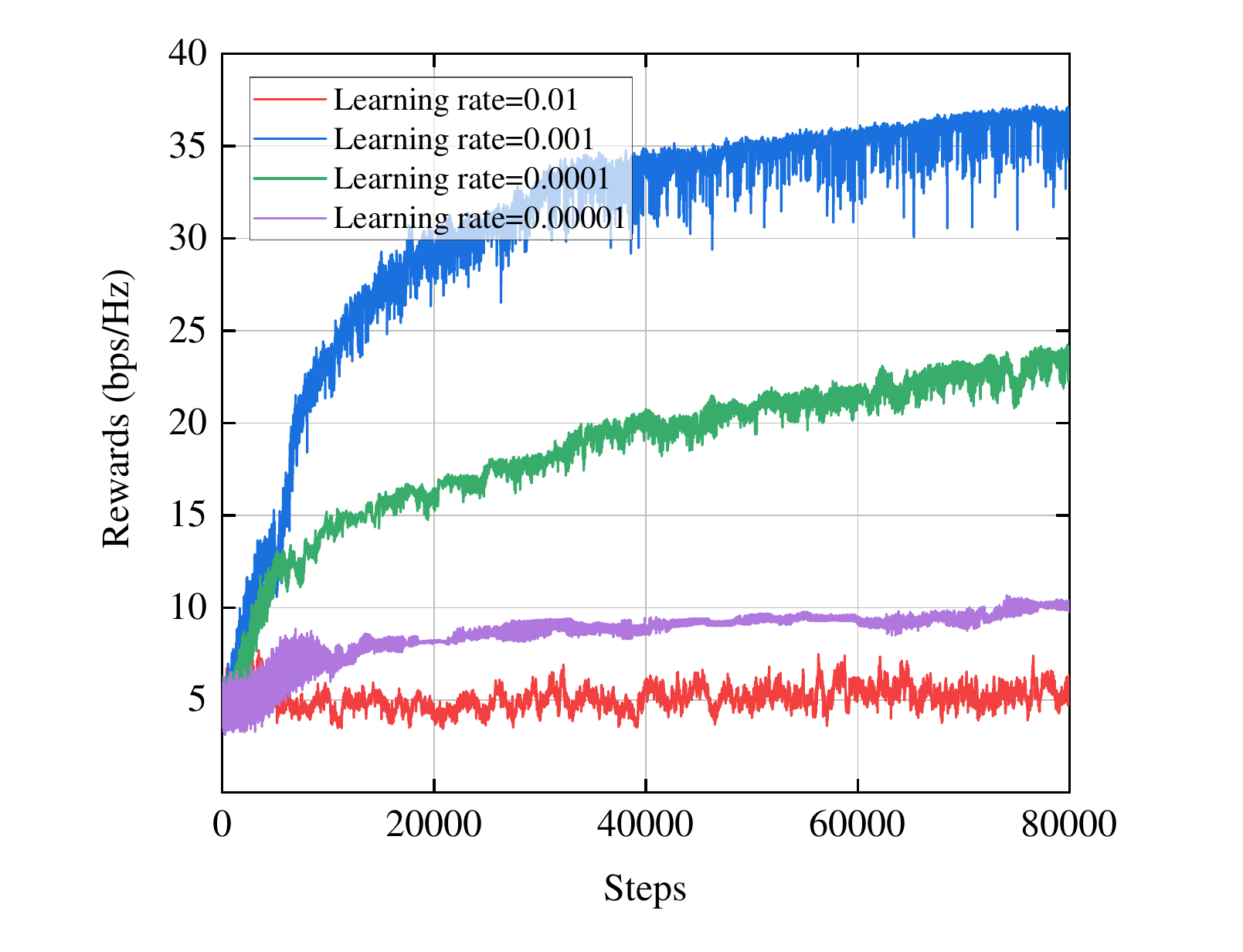}
\caption{Rewards versus time steps under different $\xi$.}
\label{fig.9}
\end{figure}

To further explore the performance of the proposed algorithm, we investigated the hyperparameters of several DRL algorithms. As shown in Fig.\ref{fig.9}, the magnitude of the learning rate $\xi^{a}=\xi^{c}$ is not positively correlated with the sum rate. When the learning rate is 0.001, the algorithm achieves the highest sum rate. It can be inferred that further adjustments to the learning rate may yield better results.

\begin{figure}[t]
\centering
\includegraphics[width=3.5in]{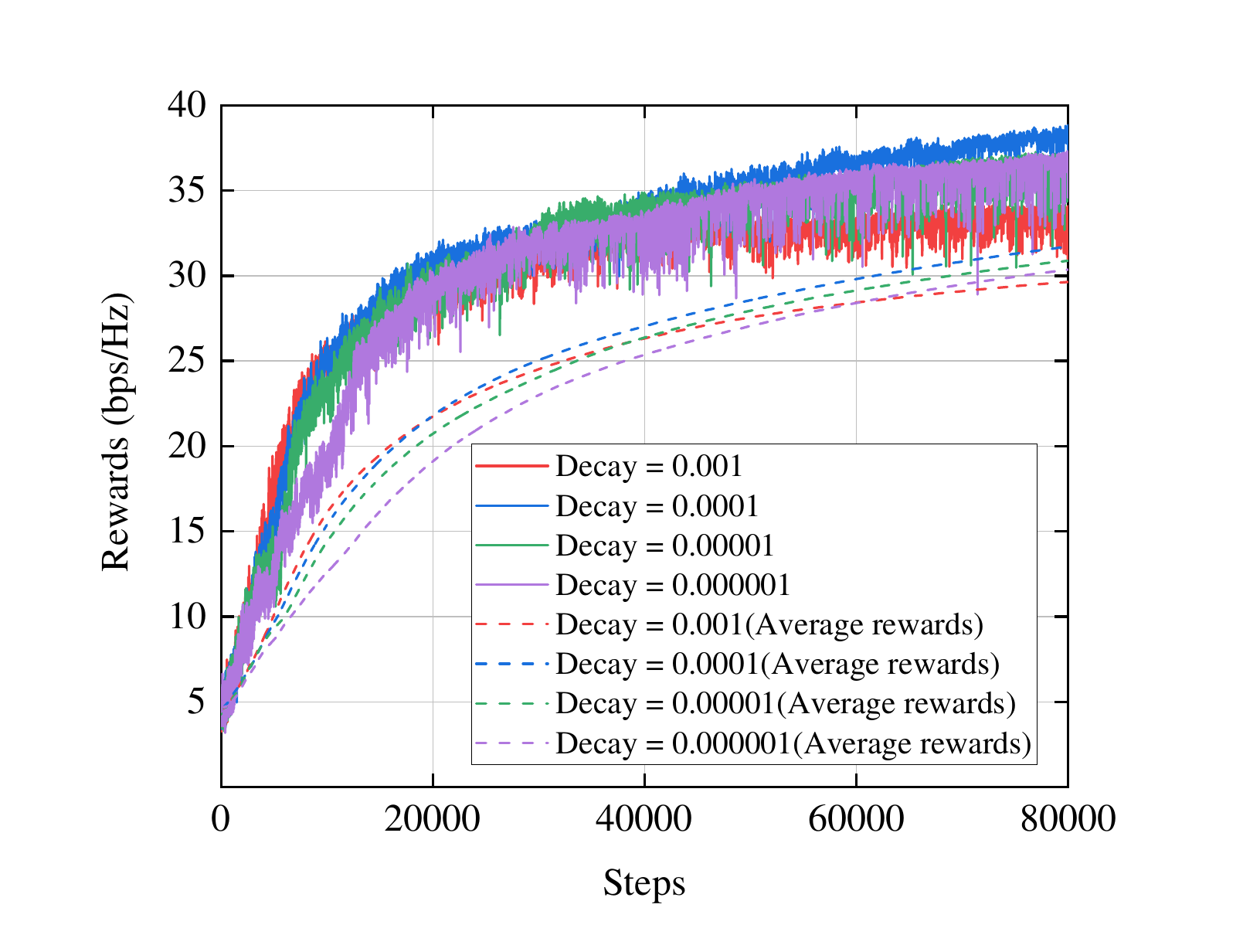}
\caption{Rewards versus time steps under different $d_c$.}
\label{fig.10}
\end{figure}

Fig.\ref{fig.10} shows the graph of the reward with time steps for different decaying rates $d_c=d_a=\{0.001,0.0001,0.00001,0.000001\}$. As shown in Fig.\ref{fig.10}, when the decaying rate is 0.0001, the reward gradually surpasses the other curves and becomes the highest as the time steps increase. This demonstrates that an appropriate decaying rate can lead to better algorithm performance. However, it is also worth noting that compared to other hyperparameters, the gaps between the four curves generated by different decaying rates are relatively small, and their convergence and stability characteristics are similar.

\begin{figure}[t]
\centering
\includegraphics[width=3.5in]{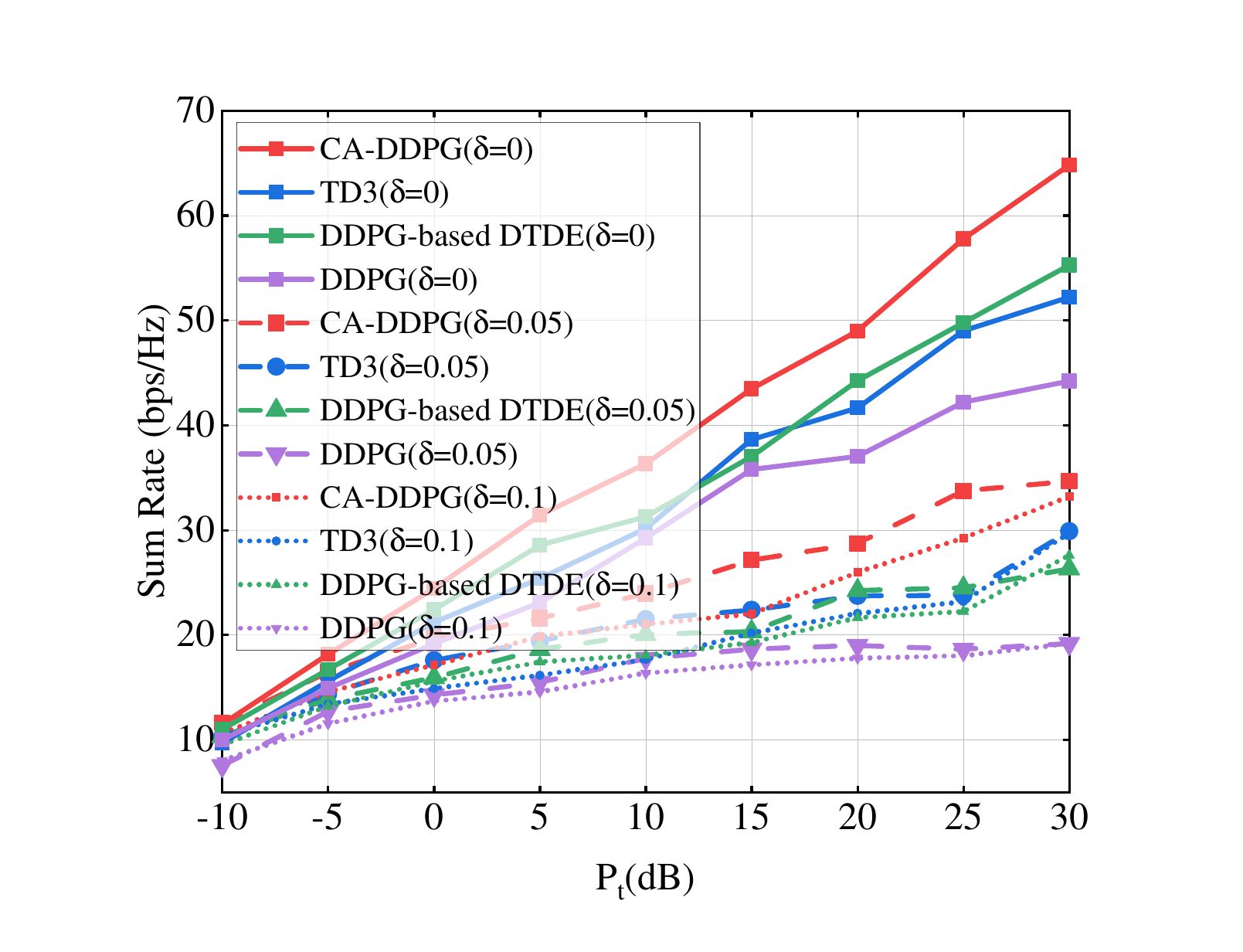}
\caption{Sum rate as a function of $P_t$ for different algorithms under different $\delta$.}
\label{fig.11}
\end{figure}

Finally, we conduct a study on the system under imperfect CSI, as shown in Fig.\ref{fig.11}. We use $\delta$ to represent the uncertainty of the channel, which is expressed as follows
\begin{equation}
\label{equ34}
\delta=\frac{\mathbb{E}\left[|\hat{h} - h|^2\right]}{\mathbb{E}\left[|\hat{h}|^2\right]} ,
\tag{34}
\end{equation}
 where $\hat{h}=h+\Delta h$ is the estimated channel, $h$ is the actual channel and $\Delta h$ is the estimated error and is assumed to follow a zero-mean Gaussian distribution. Under imperfect CSI, both our proposed algorithm and the other algorithms experience performance losses, and which increase as $\delta$ becomes larger. However, for the same $\delta$, the CA-DDPG algorithm still has a significant performance lead among the other algorithms, thus proving the feasibility of our proposed algorithm in different scenarios.

 \begin{figure}[t]
\centering
\includegraphics[width=3.5in]{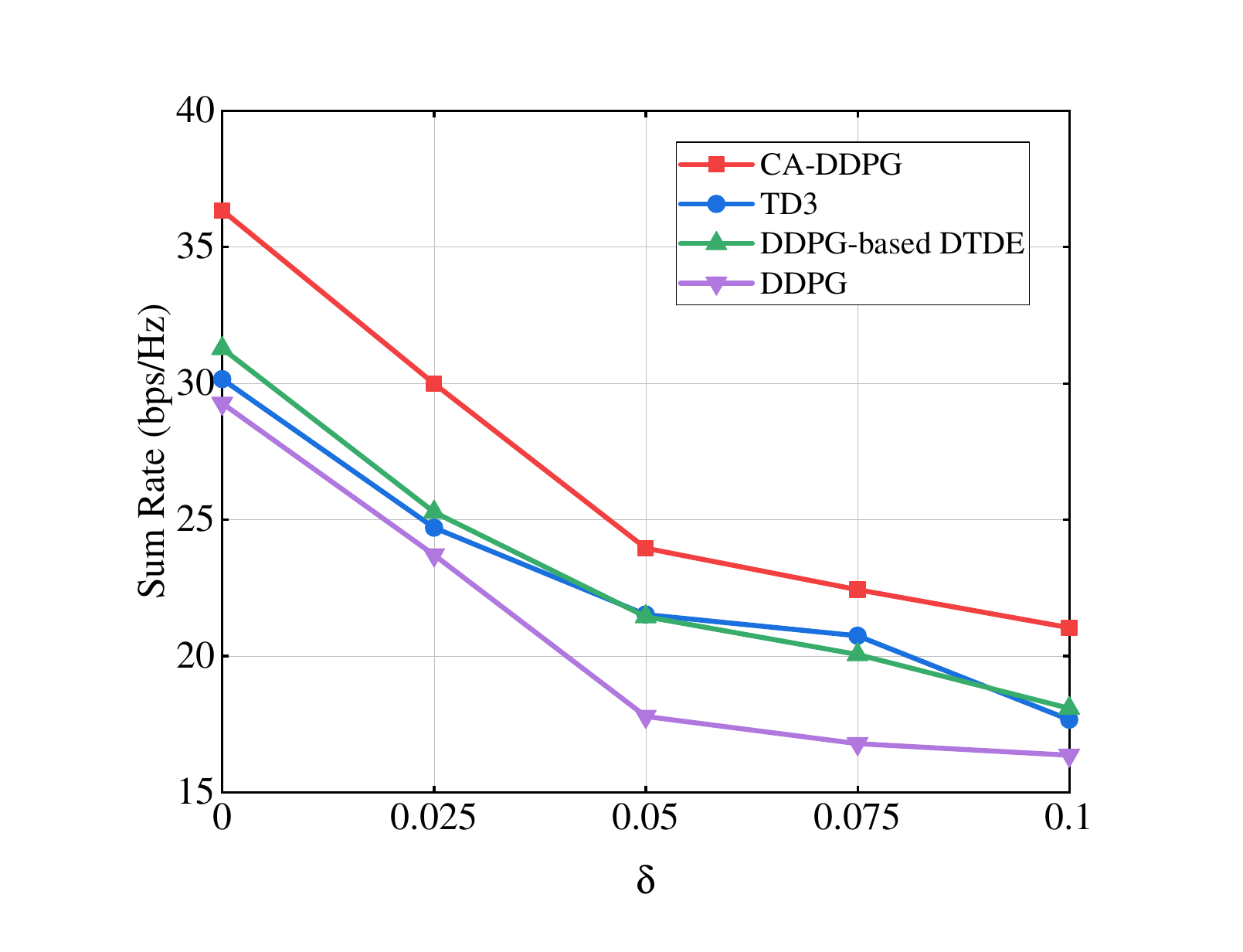}
\caption{Sum rate as a function of $\delta$ for different algorithms.}
\label{fig.12}
\end{figure}

As shown in Fig.\ref{fig.12}, as $\delta$ increases, all algorithms exhibit a certain degree of performance degradation. However, our proposed algorithm still demonstrates strong performance compared to the others, indicating that CA-DDPG can effectively improve the system's sum rate, even in harsh or dynamic channel environments.

\section{CONCLUSION}
This paper investigates the problem of beamforming, phase shift, and UAV location optimization in RIS-aided wireless communication systems. With the rapid development of modern wireless communications, the optimization problem in wireless communication has become increasingly complex and difficult to solve. To address this challenge, we leverage DRL algorithms, which have demonstrated remarkable capabilities in solving such problems. Building upon the DDPG framework, we propose the CA-DDPG algorithm to further optimize beamforming, phase shift, and UAV's location, thereby enhancing the maximum sum rate performance of RIS-aided systems. Our main innovation lies in the integration of convolutional layers into the DDPG architecture. Through convolution operations, the critic network achieves superior performance, significantly improving the maximum sum rate for users. Simulation results demonstrate that our proposed algorithm outperforms baseline methods in both low-SNR and high-SNR scenarios. Furthermore, an important extension of CA-DDPG lies in its application to multi-agent deep reinforcement learning (MADRL) frameworks. By adapting the CA-DDPG architecture to multi-agent settings, it can learn effective cooperative policies while preserving the privacy of local observations and reducing computational overhead. In addition, the proposed CA-DDPG framework demonstrates significant potential in integrated sensing and communication (ISAC) optimization scenarios. By expanding the state space of CA-DDPG to incorporate sensing-related metrics, the framework can optimize both communication and sensing performance holistically.

\bibliographystyle{IEEEtran}


\end{document}